\documentclass{article}
\usepackage[preprint]{colm2026_conference}

\usepackage{microtype}
\usepackage{xcolor}
\usepackage{hyperref}
\usepackage{url}
\usepackage{booktabs}
\usepackage{amsmath,amssymb}
\usepackage{tabularx,array}
\usepackage{enumitem}
\usepackage{lineno}
\usepackage{graphicx}
\usepackage{multirow}
\usepackage{placeins}
\usepackage{float}
\usepackage{algorithm}
\usepackage{algpseudocode}

\definecolor{darkblue}{rgb}{0,0,0.5}
\definecolor{softblue}{rgb}{0.86,0.92,0.98}
\definecolor{softgray}{rgb}{0.94,0.94,0.94}
\definecolor{softgreen}{rgb}{0.91,0.97,0.92}
\hypersetup{colorlinks=true,citecolor=darkblue,linkcolor=darkblue,urlcolor=darkblue}

\newcolumntype{Y}{>{\raggedright\arraybackslash}X}

\newcommand{\slice}[1]{\mathcal{S}(#1)}

\newcommand{\cooc}{\mathrm{co}}
\newcommand{\trgraph}{\mathrm{tr}}
\newcommand{\descop}{\operatorname{desc}}

\definecolor{editpurple}{rgb}{0.55,0.08,0.55}

\definecolor{acceptgreen}{rgb}{0.0,0.55,0.0}

\definecolor{darkblue}{rgb}{0,0,0.5}
\hypersetup{colorlinks=true,citecolor=darkblue,linkcolor=darkblue,urlcolor=darkblue}
\title{Domain-Filtered Knowledge Graphs\\
from Sparse Autoencoder Features\\
\large Contrastive filtering, multi-scale structure recovery, and edge-labeled mechanism views}
\author{John Winnicki\textsuperscript{1,*}, Abeynaya Gnanasekaran\textsuperscript{1}, Eric Darve\textsuperscript{1}\\
\textsuperscript{1}Institute for Computational and Mathematical Engineering, Stanford University, \\Stanford, CA, USA\\
\textsuperscript{*}Corresponding author: \texttt{winnicki@stanford.edu}}

\begin{document}

\ifcolmsubmission
\linenumbers
\fi

\maketitle

\begin{abstract}
%JOHN: Focus on showing novelty against the state of the art. Show that it is not incremental. 
%Next, show that the method works, and that it is improving the outcome. 
%Make sure that this is solving a problem that is important and is better. 
%Make sure it is not just "off the shelf". We developed a novel pipeline and enables things that would not be possible otherwise. 
%Key: make sure to not say this has never been done before.
Sparse autoencoders (SAEs) extract millions of interpretable features from a language model, but flat feature inventories aren't very useful on their own. Domain concepts get mixed with generic and weakly grounded features, while related ideas are scattered across many units, and there's no way to understand relationships between features. We address this by first constructing a strict domain-specific concept universe from a large SAE inventory using contrastive activations and a multi-stage filtering process. Next, we build two aligned graph views on the filtered set: a co-occurrence graph for corpus-level conceptual structure, organized at multiple levels of granularity, and a transcoder-based mechanism graph that links source-layer and target-layer features through sparse latent pathways. Automated edge labeling then turns these graph views into readable knowledge graphs rather than unlabeled layouts. In a case study on a biology textbook, these graphs recover coherent chapter and subchapter-level structure, reveal concepts that bridge neighboring topics, and transform messy sentence-level activity containing thousands of features into compact, readable views that illustrate the model's local activity. Taken together, this reframes a flat SAE inventory as an internal knowledge graph that converts feature-level interpretability into a global map of model knowledge and enables audits of reasoning faithfulness.
\end{abstract}

\section{Introduction}

Sparse autoencoders (SAEs) make it possible to represent internal language-model activations as interpretable features rather than opaque neurons \citep{bricken2023,cunningham2023,gao2024,templeton2024}. Open suites such as Gemma Scope and Llama Scope extend this approach from small-scale demonstrations to large public inventories spanning many layers and thousands of features \citep{lieberum2024gemmascope,he2024llamascope}.  

Once we move beyond flat feature lists, the problem becomes one of structure rather than enumeration. The goal shifts from adding features to characterizing how concepts are organized and used within the model. This raises three concrete questions. First, which features belong to a coherent domain-specific region, rather than generic or weakly grounded behavior? Second, within that region, how are concepts organized across a corpus, what clusters together, what separates, and what bridges neighboring topics? Third, on a given input, how do local concept neighborhoods evolve, and which upstream concepts support which downstream ones through latent mechanisms? A useful interpretability object should answer all three: it should isolate a relevant conceptual region, reveal its global organization, and expose how that structure is used in local computation.

A raw SAE inventory does not answer these questions well. Many highly active features reflect punctuation, formatting, or broad discourse patterns rather than domain-specific concepts, while semantically related ideas are often distributed across multiple neighboring features. At the corpus scale, this obscures how concepts cluster, separate, and bridge across topics. At local scale, the problem is no easier: a single sentence can activate thousands of features across the residual stream, so inspecting features one by one provides little insight into how meaningful concept neighborhoods are actually being used. To study internal conceptual structure across a corpus, we need a representation that is both selective and structured: it must isolate a coherent domain-specific region of feature space, organize the concepts within that region, and expose how local concept neighborhoods participate in downstream computation. In this work, we construct such a representation by first defining a strict domain-specific concept universe, and then organizing it with two aligned graph views: a co-occurrence graph for corpus-level structure, organized at multiple levels of granularity and a transcoder-based mechanism graph for local cross-layer explanation.

Our contributions are:
\begin{itemize}[leftmargin=1.25em,itemsep=0.2em,topsep=0.25em]
    \item We introduce \emph{a multi-stage contrastive filtering process} for building a domain-specific concept universe from a much larger SAE inventory.
    \item We build \emph{two aligned knowledge-graph views} over the retained universe: a multi-granular co-occurrence graph for corpus structure and a transcoder mechanism graph for cross-layer concept flow.
    \item We propose a \emph{hierarchical compression} over the retained universe, so dense concept and mechanism graphs become readable.
    \item We evaluate the framework on a biology textbook by testing its capability to recover \emph{chapter and subchapter-level structure, compress sentence-level activity into a readable mechanism atlas, and produce labeled edge relationships between concepts}.
\end{itemize}

\section{Related Work}

\paragraph{Sparse features at scale.}
Sparse autoencoders and related dictionary-learning methods established that many model concepts can be decomposed into sparse, often interpretable features \citep{bricken2023,cunningham2023,gao2024,templeton2024}. Open suites such as Gemma Scope and Llama Scope turn that insight into large public inventories, making it realistic to inspect many layers and many tens of thousands of features without rebuilding the sparse stack \citep{lieberum2024gemmascope,he2024llamascope}. A parallel line of work studies how those features should be described or named, from automated neuron explanations to output-centric feature descriptions \citep{bills2023neurons,gurarieh2025outputcentric}. Our paper assumes such inventories exist and asks a different question: once the inventory is large enough, what scientific object should replace a flat feature list?

\paragraph{Filtering and organization of feature spaces.}
Recent work shows that SAE feature spaces have meaningful geometry beyond isolated units. SAE geometry work studies local and global organization in feature clouds \citep{li2024geometry}, while mechanistic topic models show that interpretable features can support higher-level thematic organization \citep{zheng2025mtm}. Beyond geometry, emerging work studies graph structure directly: Clarke et al. build Jaccard-normalized same-layer co-occurrence graphs that surface strongly co-occurring communities and hub-and-spoke subgraphs, and Balcells et al. extend the graph view across adjacent layers to expose communities, pass-through motifs, and quasi-Boolean feature combinations \citep{clarke2024saecooccurrence,balcells2024evolution}. These papers are important precedents for treating feature spaces as structured objects. Our contribution differs in emphasis. We do not treat geometry alone as the endpoint; instead, we first define a strict retained universe for a specific corpus and then build multiple graph views on top of it. Thus, our approach connects structural organization of concepts to mechanistic relationships between them in a way that neither geometry nor topic models alone can provide. 

\paragraph{Circuits, transcoders, and cross-layer feature graphs.}
Mechanistic work on sparse feature circuits and transcoders moves from feature discovery toward explanatory graphs \citep{marks2024,dunefsky2024}. Circuit-tracing with transcoders further develops prompt-specific attribution graphs, supernodes, and validation tools for understanding model behavior on individual prompts \citep{ameisen2025circuit,hanna2025circuittracer}. Other work studies feature flow and matching across layers \citep{laptev2025featureflow}. Our mechanism view is closest in spirit to this literature, but the object is different. Circuit-tracing papers seek prompt-specific causal or quasi-causal graphs tied to outputs; we instead build a source SAE, a target SAE, and a transcoder into a reusable domain-filtered graph framework whose visible source-to-target edges are readable projections of latent-mediated structure.

\paragraph{Knowledge graphs and relation extraction.}
Classical knowledge graphs such as Freebase, NELL, and Wikidata represent information as entities linked by relation triples, and they motivated a large literature on knowledge-base population from text \citep{bollacker2008freebase,carlson2010nell,vrandecic2014wikidata}. That line of work spans distant supervision for relation extraction, open information extraction, and broader information-extraction pipelines that extend a graph from unstructured corpora \citep{mintz2009distant,etzioni2011openie,niklaus2018openie,martinezrodriguez2020iesemweb}. Our setting is related, but we are not extracting an external world-knowledge graph over named entities or document-level facts. Instead, we induce an internal graph over SAE features and then attach evidence-backed labels to mechanistically induced edges. Thus, relation-extraction work mainly informs our edge-labeling layer, while the node universe and candidate edges come from activation structure and transcoder-mediated feature flow (rather than from entity linking and sentence-level triple extraction).

\section{From a Flat SAE Inventory to a Strict Concept Universe}
\label{sec:filter}

This section describes how we turn a large SAE feature inventory into the semantic node set used by the rest of the paper. The procedure has two stages. First, we use activation statistics on the target corpus and contrast corpora to form an inclusive shortlist of features that have enough support, enrichment, or localization to merit closer inspection. Second, we convert each shortlisted feature into an evidence packet and determine whether the feature corresponds to a visible, domain-relevant, and contrast-distinctive concept. The result is not yet a graph. It is a strict retained concept universe that fixes the nodes for all later co-occurrence and mechanism views.

\subsection{Why graph construction starts with filtering}

Even when restricting to a domain-focused corpus, many active SAE features track punctuation, discourse management, formatting fragments, broad quantitative language, or other weakly grounded patterns. If all such features enter the graph, the result reflects generic model behavior as much as the conceptual structure of the target corpus. Filtering is therefore the first modeling step rather than a cleanup pass after graph construction.

Let $\mathcal{V}$ denote the candidate inventory and let $\mathcal{V}^{\star} \subset \mathcal{V}$ denote the retained strict concept universe. All downstream graphs in this paper are built on $\mathcal{V}^{\star}$ rather than on the raw inventory $\mathcal{V}$.

\subsection{A contrastive shortlist from activation evidence}

The first pass constructs a high-recall candidate set. Contrast is useful because many undesirable features are active in many corpora: they are present, but not specific. We therefore compare the target corpus against contrast corpora and keep candidates that look promising along at least one of three axes: \emph{support} (is the feature meaningfully present in the target corpus?), \emph{enrichment} (is it more characteristic of the target than of the contrasts?), and \emph{localization} (does it concentrate in coherent chapter or subchapter regions rather than diffusing everywhere?). These statistics are meant to remove obvious boilerplate such as grammar and syntax features while preserving recall; they do not by themselves decide semantic relevance. The exact thresholds and ranking details are deferred to Appendix~\ref{app:filter-details}.

\subsection{Packet-level adjudication}

Shortlist statistics are intentionally coarse, so each surviving feature is turned into an auditable evidence packet. A packet combines the description hypothesis, representative target windows, contrast windows, local chapter and subchapter context, and shortlist tags. The staged adjudicator then asks three feature-level questions: is the proposed concept \emph{visible} in the target evidence, does it \emph{belong here} as part of the target-domain region rather than as incidental background, and is that fit \emph{distinctive} relative to the contrast evidence? Appendix~\ref{app:filter-details} gives the packet schema and decision map. The output of the filtering stage is therefore the strict universe $\mathcal{V}^{\star}$: a retained set of readable, target-relevant SAE features, together with their evidence packets and adjudication decisions. This stage fixes the node set for the rest of the paper. The next section keeps this universe fixed and constructs edge-labeled graph views over it: corpus-scale co-occurrence graphs, local transcoder mechanism graphs, hierarchy-aware compressed views, and evidence-backed edge labels.

\section{Edge-Labeled Knowledge Graph Construction}
\label{sec:graphs}

The filtering stage defines the node universe, while the graph pipeline determines what can be learned once that universe is fixed. We use two complementary views over the same retained concept set. The co-occurrence graph is the corpus-scale view: it asks which concepts recur together across sentences, paragraphs, subchapters, and chapters. The transcoder graph is the local mechanism view: it asks which readable source-layer concepts appear to support which readable target-layer concepts on a selected input. We construct an abstraction hierarchy to organize the retained universe, supporting optional grouped subviews inside that universe, and later compressing dense local mechanism graphs into readable supernode views. Appendix~\ref{app:methods-detail} retains the full hierarchy construction, optional hierarchy-defined subview semantics, the full static/dynamic mechanism derivation, and the hierarchy-respecting compression rule, together with Figures~\ref{fig:abstraction-tree}, \ref{fig:transcoder_overview}, and \ref{fig:transcoder_compression}. Appendix~\ref{app:workflow-detail} details the full auto-relate workflows together with the cross-view operational examples.

% \subsection{A retained concept universe, reused everywhere}

% The crucial design decision is to reuse one retained semantic universe across every downstream view. If $\mathcal{V}^{\star}$ is the strict biology concept universe produced by Section~\ref{sec:filter}, then every later graph is built inside $\mathcal{V}^{\star}$ rather than on the full SAE inventory. This is what makes the views comparable. The chapter and subchapter density figures in the main text, the local sentence mechanisms, and the edge-labeled case studies all refer back to the same filtered node set rather than to separate ad hoc slices.

% A hierarchy is still built over this retained universe, but its function has changed relative to the older version of the paper. It is no longer the first object the paper asks the reader to understand. Instead, it provides semantic organization, optional grouped views within the retained universe, and most importantly the compression machinery used when local mechanism graphs become too dense to read comfortably. 

\subsection{Co-occurrence graphs across textbook granularity}

The first graph family answers a corpus question: once a node belongs to the retained universe, what does it tend to appear with? Presence is defined at the sentence level and then lifted deterministically to larger units. Let $h_{i,v}$ denote the activation of feature $v \in \mathcal{V}^{\star}$ on token $i$, and let $h^+_{i,v}=\max(h_{i,v},0)$. For sentence $s$, we score feature $v$ by its strongest positive token activation,
\[
 m_{s,v}=\max_{i\in s} h^+_{i,v},
 \qquad
 X^{(\mathrm{sent})}_{s,v}=\mathbf{1}[m_{s,v}>\theta_v],
\]
where $\theta_v$ is a feature-specific threshold. Larger textual units are then defined from sentence presence alone. For granularity $g \in \{\mathrm{paragraph},\mathrm{subchapter},\mathrm{chapter}\}$ and unit $u$ with constituent sentences $\mathcal{C}_g(u)$,
\[
X^{(g)}_{u,v}=\mathbf{1}[\exists s\in \mathcal{C}_g(u): X^{(\mathrm{sent})}_{s,v}=1].
\]
Co-occurrence counts are built from the resulting binary presence matrices,
\[
C^{(g)}=(X^{(g)})^\top X^{(g)},
\qquad
J^{(g)}_{ab}=\frac{C^{(g)}_{ab}}{C^{(g)}_{aa}+C^{(g)}_{bb}-C^{(g)}_{ab}},
\]
and sparse same-layer graphs are obtained by retaining strong neighbors per node and symmetrizing. Appendix~\ref{app:methods-detail} gives the fuller notation that also supports optional hierarchy-defined grouped subviews inside $\mathcal{V}^{\star}$.

Three points are worth emphasizing. First, sentence presence is the primitive event. Paragraph, subchapter, and chapter graphs are lifted views of the same set of sentences. Second, because filtering happens first, these co-occurrence graphs are not polluted by the full inventory's punctuation-like or boilerplate-heavy regions. Third, the co-occurrence view is not only about neighborhoods: it is also where bridge concepts become visible. Dense chapter-specific regions, transitional corridors, and subchapter basins are all properties of the same retained graph and are easiest to see in the density figures and bridge examples later in the paper.

\subsection{Transcoder mechanism graph}

The co-occurrence graph tells us what retained concepts tend to go together across the corpus. The transcoder graph answers a different question: on a selected unit $q$, which readable source-layer concepts appear to support which readable target-layer concepts through sparse latent mechanisms? We place a source SAE, a target SAE, and a transcoder on the same layer transition, so that the transcoder remains the mechanism object while the SAE dictionaries make that mechanism readable.

At a high level, each visible source$\rightarrow$target feature edge in the readable graph is a projection of one or more sparse latents that read from the source-side residual stream and write to the target-side residual stream. Because the SAE dictionaries make those directions readable, we can express the resulting mechanism as source-side features writing into target-side features. The full residual-space derivation, the source and target-support matrices, the static mechanism captions, and the dynamic edge-evidence equations are detailed in Appendix~\ref{app:methods-detail}. The key object is the unit-conditioned directed graph $G^{\trgraph}(q)$ obtained by combining three kinds of evidence: source-feature activity, latent activity, and target-feature activity. A displayed edge in this formulation is therefore not merely a topic association. It is a readable summary of a local cross-layer pathway that is supported by actual sentence evidence on the chosen unit.

We distinguish two graph objects: a static graph and a dynamic graph. Here a unit $q$ refers to the specific text span being audited, such as a sentence, paragraph, or other corpus segment for which we have source-feature, latent, and target-feature activations. In the static graph, the transcoder together with the SAE dictionaries defines a library of possible readable pathways based on summary activation statistics. In a dynamic graph for a selected unit $q$, the chosen text span determines which of those pathways are actually active. The transcoder graphs shown in the results are therefore drawn from dynamic, unit-conditioned evidence rather than from a prompt-independent summary alone.

\subsection{Hierarchy-aware compression}

Even after filtering, local mechanism graphs can be too dense for a human audit. This is where the abstraction hierarchy becomes operationally useful by compressing the dynamic graph. We accomplish this by collapsing coherent regions of the precomputed abstraction hierarchy into supernodes with grounded higher-level labels. Internal hierarchy nodes summarize coherent groups of retained features: when a local graph is too dense, those groups can collapse into supernodes while preserving the strongest cross-group traffic and blocking collapse across internally important active edges. The result is a mixed view of leaves and supernodes that remains faithful to the active local graph while being much more readable. Appendix~\ref{app:methods-detail} gives the mixed-cover rule, the blocking-edge construction, and the full compression figure.

\subsection{Edge semantics and why the object is a knowledge graph}

Weighted edges alone are not enough. To turn the co-occurrence and mechanism views into actual knowledge graphs, the edges must carry readable semantics backed by evidence. This labeling step is analogous to Neuronpedia-style autointerpretation~\citep{lin2023neuronpedia,bills2023language}, but applied to edges rather than individual features: instead of assigning a readable description to a latent from its activation evidence, we assign a readable relation to an edge from the evidence supporting that connection. Co-occurrence edges are labeled from joint and contrastive evidence sentences, while directed mechanism edges use the strongest sentence-level evidence together with latent-level hints to propose readable relation phrases. The labeling workflows share the same principle: labels should describe what the edge means in evidence, not merely restate the endpoint feature names.

The co-occurrence graph tells us which concepts recur together and where bridge structure appears; the transcoder graph tells us how a local concept neighborhood is transformed across layers; and the auto-related edge labels make both views readable as knowledge graphs rather than as weighted node maps. Appendix~\ref{app:workflow-detail} gives the fuller edge-labeling story.

\section{Experimental Setting}
\label{sec:setup}

\subsection{Target corpus, contrast corpora, and sparse stack}

Our target corpus is a sentence-segmented biology textbook corpus derived from \emph{Concepts of Biology}~\citep{molnar2015conceptsbiology}. For contrastive filtering we use \emph{Modern World History}~\citep{allosso2021modernworldhistory} and \emph{Physical Geology}~\citep{earle2015physicalgeology}. These contrasts are broad expository textbooks outside biology, which helps separate domain-specific biology concepts from generic textbook or nonfiction features. Sentences are grouped into paragraphs, subchapters, and chapters. The sparse stack consists of Gemma Scope residual SAEs at layers 20 and 21 (width 65k), together with a layer-20 width-16k transcoder that maps the layer-20 residual stream to layer 21. From this stack we construct the two graph views: a concept-level co-occurrence graph and a mechanism-level transcoder graph.

\subsection{Evaluation focus}

Our evaluation asks two questions. First, do the filtered graphs recover textbook organization across scales on shared coordinates? Second, after filtering and compression, do local sentence-level graphs remain readable and relevant to the selected sentence? We begin with density maps on shared sentence-graph coordinates because they give the clearest view of structure recovery without confounding the result with separate layouts; density-estimation views make localized structure easier to inspect, and scalable implementations make such maps practical on large point clouds \citep{epstein2025sdkde,epstein2026flashsdkde}. Figure~\ref{fig:structure-main} combines representative chapter views with a respiratory-system drill-down, while Appendix Figure~\ref{fig:chapter-density-full} gives the full chapter-density profile to confirm that activations within a chapter (and subchapter) have highly localized activation patterns within our graph structure. We then trace the local progression from the full active sentence feature set to a sentence-induced co-occurrence graph, a compact active concept neighborhood, and finally a compressed sentence-level transcoder view. This progression separates the roles of selection, local relational structure, abstraction, and mechanism. We close with a case study on our knowledge graphs that examines a cross-topic co-occurrence structure and a directed sentence-level mechanism graph.

\section{Results}
\label{sec:results}

Flat SAE feature lists are hard to inspect at corpus scale, so the main empirical question is whether graph structure recovers a coherent organization of the underlying activations. This is the key evaluation because flat SAE inventories mix domain-relevant and generic units, scatter related ideas across many features, and do not explicitly represent relations~\citep{cunningham2023sparse,bricken2023monosemanticity,templeton2024scaling}. Recovering textbook structure would therefore show that the graph is organizing concepts, not merely reformatting a feature list. We first evaluate multiscale structure recovery. Biology textbooks provide a useful external reference because chapters and subchapters encode a human-authored topical organization. Those labels are not used as supervised targets for graph layout or clustering, although chapter/subchapter segmentation and localization statistics do enter the filtering and packet-adjudication pipeline. If the induced graphs still place material from the same chapter in nearby regions, preserve finer subchapter refinement within a chapter basin, and expose bridge structure between related regions, then they are capturing organization that goes beyond a visually tidy layout.

This is still a nontrivial test: the graph is not optimized to reproduce chapter boundaries, biology chapters share substantial vocabulary, and many SAE features are generic or only weakly informative~\citep{karvonen2025saebench}. A graph built only from local activations could easily organize around boilerplate language, blur into one broad biology cluster, or produce layouts that look coherent without actually tracking topic structure. Recovering textbook structure, therefore, matters for more than presentation. It is evidence that filtering, co-occurrence, and compression are isolating a usable map of the model's internal organization of the domain. 

We then evaluate local sentence-level interpretability. After filtering and compression, a good local graph should retain the features that matter for understanding a sentence, show which concepts connect nearby regions in the co-occurrence view, and expose plausible directed interactions in the transcoder view.

\subsection{Shared-coordinate activation maps recover textbook structure at multiple scales}

\begin{figure}[t]
    \centering
    \includegraphics[width=\textwidth]{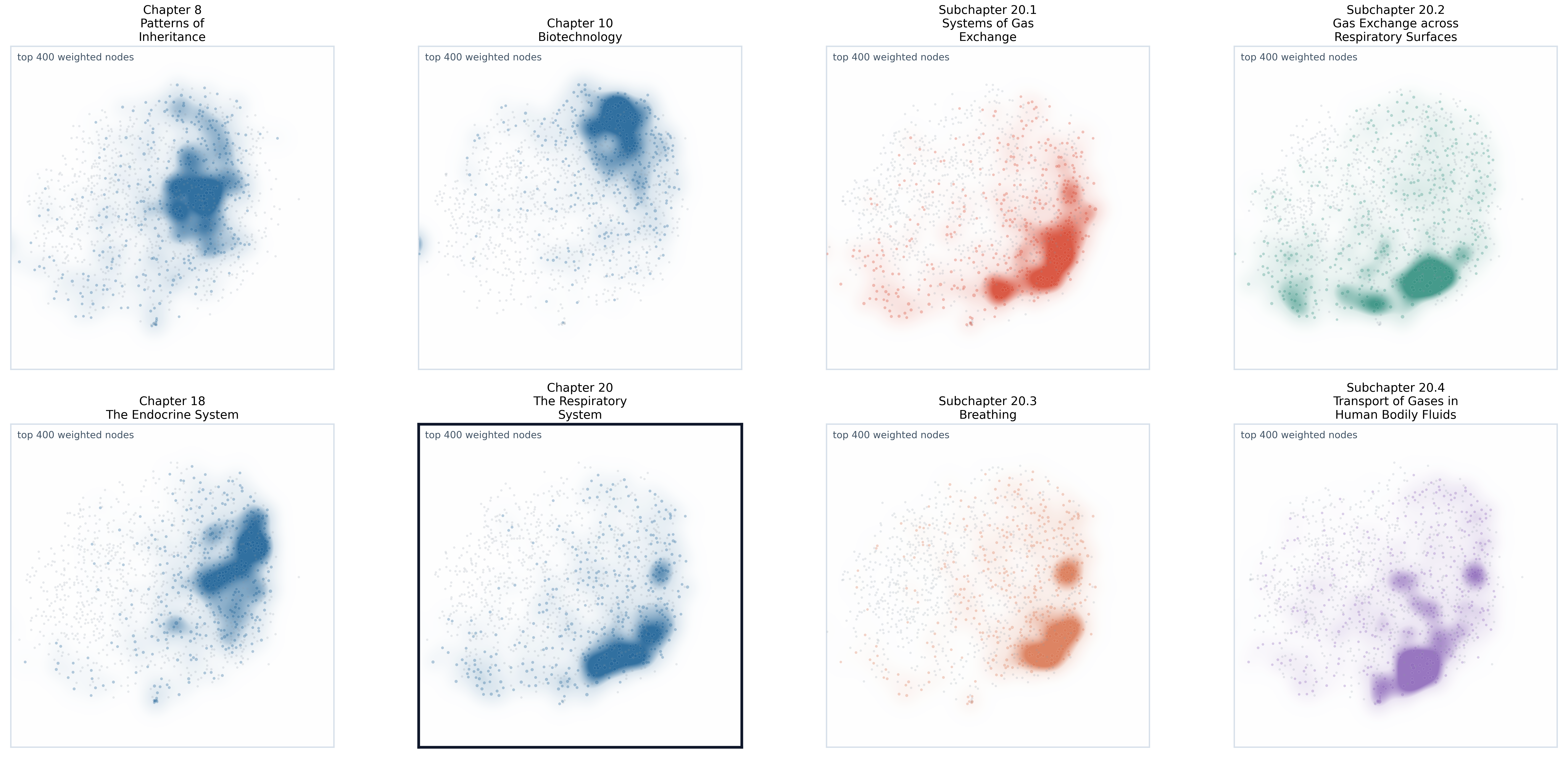}
    \caption{Shared-coordinate activation maps recover textbook structure at multiple scales. The chapter panels show distinct global basins for representative chapters, while the respiratory-system subchapter panels show overlapping but still distinguishable local modes inside a shared chapter region.}
    \label{fig:structure-main}
\end{figure}

We begin with density maps on shared sentence-graph coordinates because they give the clearest view of structure recovery. Figure~\ref{fig:structure-main} visualizes the sentence-level co-occurrence graph with the SAE activation heatmap overlaid. Every panel is a weighted view of the same shared sentence-graph coordinates, so the differences across panels cannot be explained away by separate layouts. The chapter panels show compact high-density basins in different regions of the retained universe rather than one diffuse biology cloud: inheritance, biotechnology, endocrine-system content, and respiratory-system content all place their activation mass in distinct neighborhoods. This is exactly the kind of structure flat feature lists fail to provide: related concepts consolidate into stable regions of a common space rather than remaining dispersed across many partially informative units \citep{li2024geometry,zheng2025mtm}.

The same figure also probes whether the graph preserves finer structure within a single chapter region. The four respiratory subchapters are shown as co-occurrence heatmaps on the same graph, with intensity determined by SAE feature activation weight. They overlap within a common respiratory neighborhood, as expected, but they are not interchangeable: each places its strongest density mass on a distinct mode within that broader neighborhood. This suggests that the graph captures not only chapter-level separation, but also finer within-chapter organization.

\FloatBarrier

This matters because the graph is not directly trained to recover chapter or subchapter labels, yet still aligns with them across scales. Figure~\ref{fig:structure-main} is therefore not just a visualization sanity check. It is an external test of whether the selected features and their induced relations recover coherent domain organization present in both the corpus and the model’s internal representation of it. Appendix Figure~\ref{fig:chapter-density-full} reinforces the same point at full-book scale by showing the complete chapter-density atlas. Appendix Table~\ref{tab:structure-metrics} provides a quantitative complement to these shared-coordinate maps. The strongest structure-recovery scores occur at paragraph and sentence aggregation, where chapter and subchapter alignment are both high, same-chapter edge mass is concentrated, and within-chapter nodes lie closer to one another than nodes from different chapters. This matches the qualitative picture in Figure~\ref{fig:structure-main}: the graph is not only visually organized, but measurably aligned with textbook structure at multiple scales.

\subsection{From diffuse sentence activity to a readable mechanism atlas}

\begin{figure}[t]
    \centering
    \includegraphics[width=\textwidth]{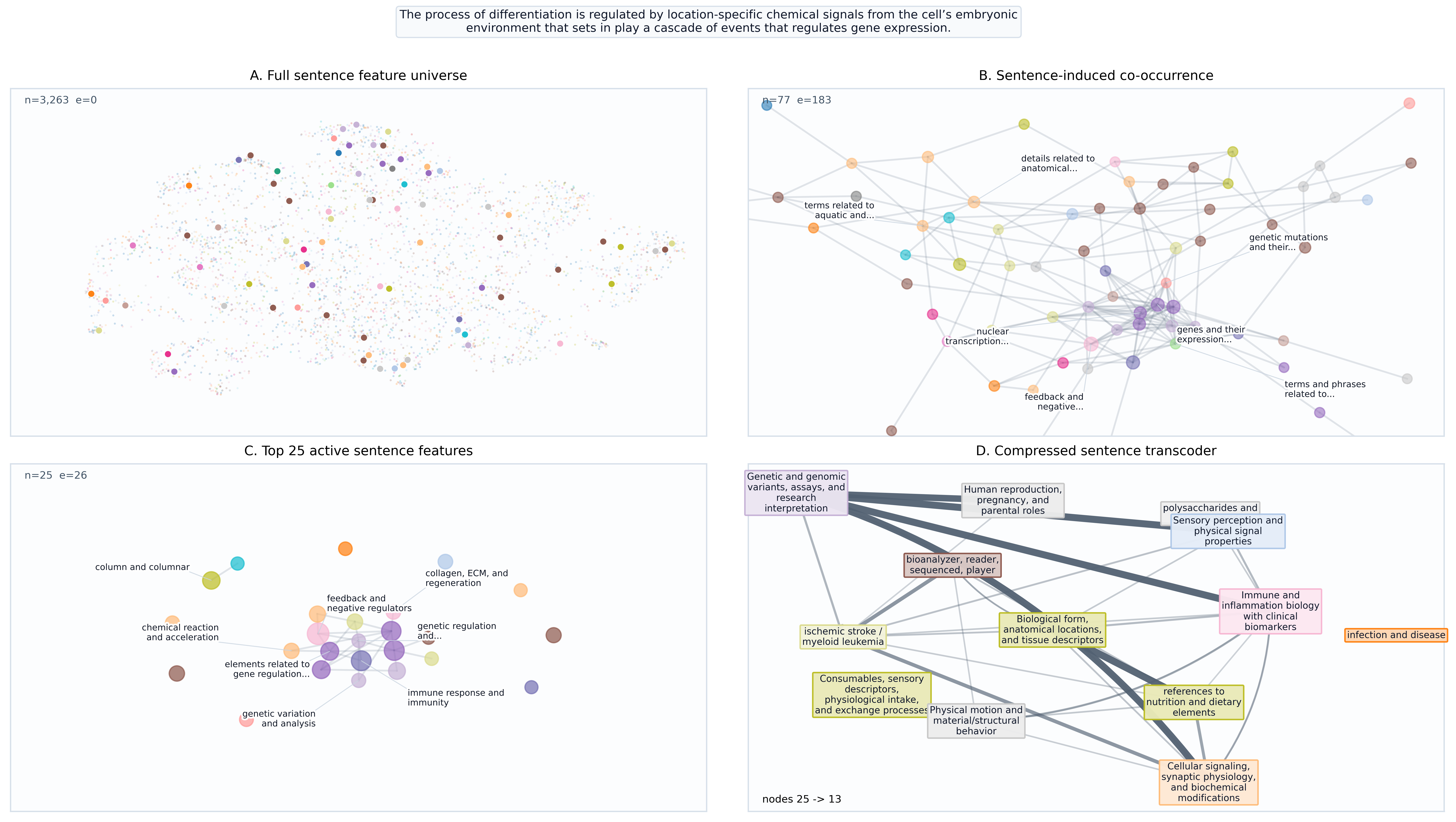}
    \caption{Capability progression from diffuse sentence activity to a readable mechanism atlas. The example moves from full sentence activity to a sentence-induced co-occurrence graph, a compact top-feature neighborhood, and a compressed sentence-level transcoder view.}
    \label{fig:progression}
\end{figure}

The density maps establish global and local structure recovery. Figure~\ref{fig:progression} asks a complementary question: once a sentence is selected, how does the framework turn the full set of active sentence features into a readable explanatory object? Panel~A starts from that full feature set, which is too diffuse to function as an explanation on its own. Panel~B induces a sentence-level co-occurrence graph over the activated concepts and already yields a more legible local map. Panel~C trims the view to the top active sentence features, exposing the tight concept neighborhood that dominates the sentence. Panel~D then compresses the sentence-level transcoder graph into a mechanism atlas whose supernodes summarize the strongest local pathways.

The progression matters because each stage exposes a different level of structure in the same sentence-level activity. Panel A is diffuse and difficult to interpret. Panel B introduces local relational structure among co-active concepts \citep{clarke2024saecooccurrence,balcells2024evolution}. Panel C isolates the dominant concept neighborhood. Panel D then maps this neighborhood into mechanism space, yielding a directed representation of interactions between concepts \citep{dunefsky2024,ameisen2025circuit,hanna2025circuittracer}.

The result is not simply a smaller view, but a more structured one: a compact representation that captures both the local organization of concepts and their downstream interactions. More generally, dimensionality reduction and low-dimensional projections are often what make large high-dimensional measurement or representation spaces inspectable in the first place, turning raw measurements or learned representations into objects that can be organized, clustered, and explored \citep{winnicki2024matrix,winnicki2025lzwaveforms,jolliffe2016pca,vandermaaten2008tsne,mcinnes2018umap}.

\FloatBarrier

\subsection{Auto-related edges complete the move from layout to knowledge graph}

\begin{figure}[t]
    \centering
    \includegraphics[width=\textwidth]{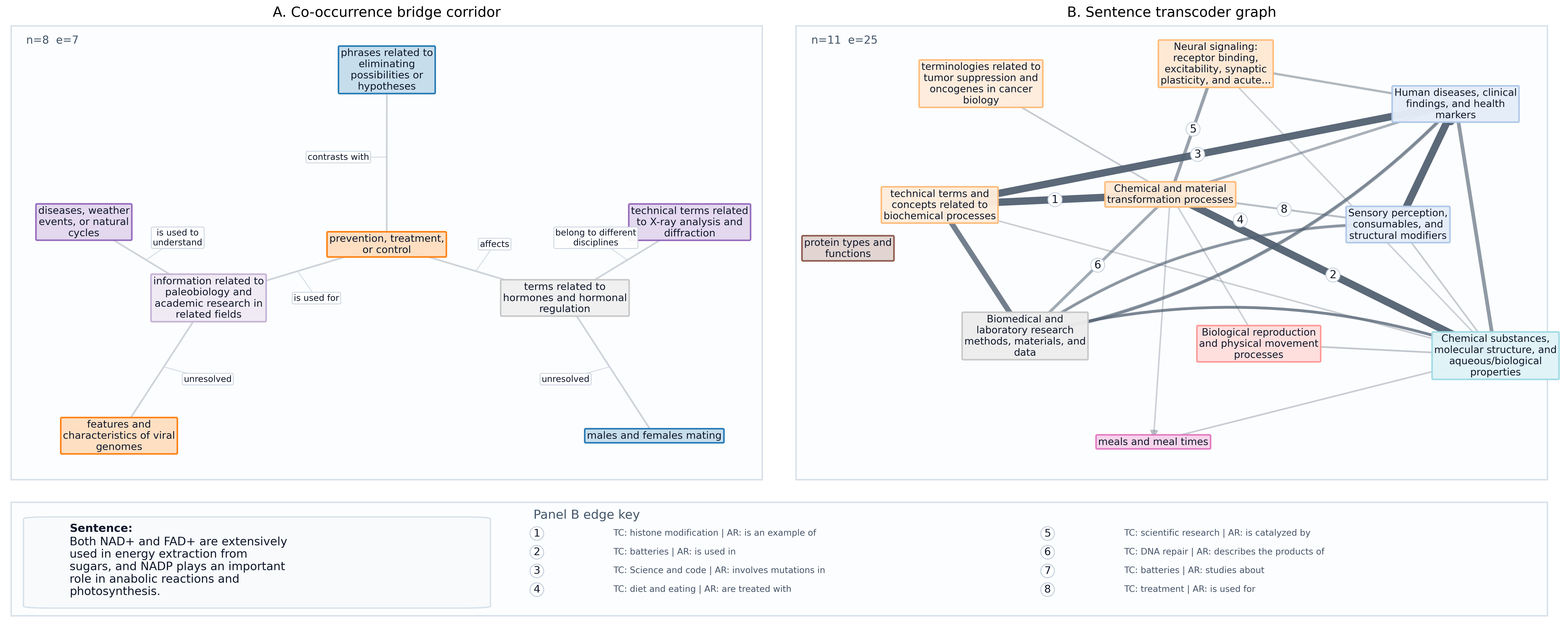}
    \caption{Edge-labeled knowledge-graph views. Panel A shows a co-occurrence bridge corridor with readable relation phrases on the edges. Panel B shows the companion sentence-level transcoder graph with numbered directed edges. The legend demarcates the transcoder (TC) neuronpedia edges and the auto-relate (AR) generated edge labels.}
    \label{fig:edgekg}
\end{figure}

Figure~\ref{fig:edgekg} visualizes the final knowledge graph produced by the pipeline. Panel A extracts a local bridge corridor from the co-occurrence graph and shows that its edges admit readable relation labels. Panel B shows the corresponding sentence-level transcoder graph with directed edges, where each edge is paired with a mechanism label and its associated auto-relate description. This figure highlights the endpoint of the method. The goal is not only to identify a relevant set of features or a coherent local neighborhood, but to recover a graph whose edges are semantically interpretable. At the sentence level, the object is therefore not a collection of activations, but a structured representation in which concepts and their interactions can be directly inspected. This is important because flat feature sets do not encode relations, and local activation maps alone do not specify how concepts connect. The labeled graph closes this gap by making both concept-level structure and mechanism-level interactions explicit within a single, local object.

\FloatBarrier

\section{Discussion and Future Work}

After strict domain filtering, the retained features recover textbook organization at multiple scales, transform messy sentence activity into a readable local view, and support edge-labeled graphs at both the co-occurrence and transcoder levels. Taken together, these results show that SAE features can be organized into a structured internal knowledge graph, rather than left as a flat inventory.

The larger motivation is to investigate whether activation-derived concept structure can provide a  more faithful view of reasoning than generated chain-of-thought \citep{turpin2023unfaithful,lanham2023faithfulness}. We do not claim that result here. Instead, this paper establishes the prerequisite object for asking it: a filtered, multi-scale, edge-bearing internal concept graph that can later be compared against generated rationales, faulty reasoning traces, confidence reports \citep{wang2023pinto,abdollahi2025demystifying,epstein2025fermieval}, or other external explanations.

These graph objects can also be explored through an interactive browser artifact located at \url{https://saegraph-render-dual-private.onrender.com/} that exposes the retained concept universe, local edge labels, and compressed mechanism views at full corpus scale. This makes it possible to inspect neighborhoods beyond the static examples in Figures~\ref{fig:progression} and \ref{fig:edgekg}, trace additional bridge regions, and examine sentence-level mechanism views for other inputs. In this sense, the browser artifact extends the paper's qualitative analysis by making the learned graph structure directly navigable.

The most important methodological extension is to  supplement the current projected transcoder edge weights with attribution-derived edges. Natural next steps include attention-mediated attribution through transcoders and recursive graph tracing \citep{dunefsky2024,ameisen2025circuit}, reused for both local dynamic graphs and the static atlas. More broadly, the same framework can be extended beyond a single layer pair to trace multi-step pathways across larger portions of the network. The point of this submission is not that the method is confined to two layers, but that even a minimal two-layer demonstration already yields a usable internal knowledge-graph object.

\section{Conclusion}

The right starting point for SAE knowledge graphs is not a hierarchy-first workflow over the full inventory, but a strict domain filter that fixes a usable concept universe. Once that universe is fixed, density maps on shared graph coordinates show that the retained features recover textbook organization from chapters down to subchapters. Local mechanism views show how that same universe can be turned into directed, readable sentence-scale explanations after hierarchy-aware compression. The resulting object is an internal knowledge graph over model concepts rather than a flat feature list. That object, in turn, sets up the next scientific question: whether activation-derived concept graphs can explain model reasoning more faithfully than generated rationales.

\bibliographystyle{colm2026_conference}
\bibliography{references}

@article{bricken2023,
  title   = {Towards Monosemanticity: Decomposing Language Models with Dictionary Learning},
  author  = {Bricken, Trenton and Templeton, Adly and Batson, Joshua and Chen, Brian and Jermyn, Adam and Conerly, Tom and Turner, Nicholas L. and others},
  journal = {Transformer Circuits Thread},
  year    = {2023},
  url     = {https://transformer-circuits.pub/2023/monosemantic-features/index.html}
}

@inproceedings{cunningham2023,
  title     = {Sparse Autoencoders Find Highly Interpretable Features in Language Models},
  author    = {Huben, Robert and Cunningham, Hoagy and Riggs Smith, Logan and Ewart, Aidan and Sharkey, Lee},
  booktitle = {International Conference on Learning Representations},
  year      = {2024},
  url       = {https://openreview.net/forum?id=F76bwRSLeK}
}

@inproceedings{gao2024,
  title     = {Scaling and Evaluating Sparse Autoencoders},
  author    = {Gao, Leo and Dupr{\'e} la Tour, Tom and Tillman, Henk and Goh, Gabriel and Troll, Rajan and Radford, Alec and Sutskever, Ilya and Leike, Jan and Wu, Jeffrey},
  booktitle = {International Conference on Learning Representations},
  year      = {2025},
  url       = {https://openreview.net/forum?id=tcsZt9ZNKD}
}

@article{templeton2024,
  title   = {Scaling Monosemanticity: Extracting Interpretable Features from Claude 3 Sonnet},
  author  = {Templeton, Adly and Bricken, Trenton and Batson, Joshua and Chen, Brian and Jermyn, Adam and Conerly, Tom and others},
  journal = {Transformer Circuits Thread},
  year    = {2024},
  url     = {https://transformer-circuits.pub/2024/scaling-monosemanticity/}
}

@inproceedings{lieberum2024gemmascope,
  title     = {Gemma Scope: Open Sparse Autoencoders Everywhere All At Once on Gemma 2},
  author    = {Lieberum, Tom and Rajamanoharan, Senthooran and Conmy, Arthur and Smith, Lewis and Sonnerat, Nicolas and Varma, Vikrant and Kram{\'a}r, J{\'a}nos and Dragan, Anca and Shah, Rohin and Nanda, Neel},
  booktitle = {Proceedings of the 7th BlackboxNLP Workshop: Analyzing and Interpreting Neural Networks for NLP},
  year      = {2024},
  url       = {https://aclanthology.org/2024.blackboxnlp-1.19/}
}

@article{he2024llamascope,
  title   = {Llama Scope: Extracting Millions of Features from Llama-3.1-8B with Sparse Autoencoders},
  author  = {He, Zhengfu and Shu, Wentao and Ge, Xuyang and Chen, Lingjie and Wang, Junxuan and Zhou, Yunhua and Liu, Frances and Guo, Qipeng and Huang, Xuanjing and Wu, Zuxuan and Jiang, Yu-Gang and Qiu, Xipeng},
  journal = {arXiv preprint arXiv:2410.20526},
  year    = {2024},
  url     = {https://arxiv.org/abs/2410.20526}
}

@article{bills2023neurons,
  title   = {Language Models Can Explain Neurons in Language Models},
  author  = {Bills, Steven and Cammarata, Nick and Mossing, Dan and Tillman, Henk and Gao, Leo and Goh, Gabriel and Sutskever, Ilya and Leike, Jan and Wu, Jeffrey and Saunders, William},
  journal = {OpenAI},
  year    = {2023},
  url     = {https://openaipublic.blob.core.windows.net/neuron-explainer/paper/index.html}
}

@inproceedings{gurarieh2025outputcentric,
  title     = {Enhancing Automated Interpretability with Output-Centric Feature Descriptions},
  author    = {Gur-Arieh, Yoav and Mayan, Roy and Agassy, Chen and Geiger, Atticus and Geva, Mor},
  booktitle = {Proceedings of the 63rd Annual Meeting of the Association for Computational Linguistics (Volume 1: Long Papers)},
  year      = {2025},
  pages     = {5757--5778},
  url       = {https://aclanthology.org/2025.acl-long.288/},
  doi       = {10.18653/v1/2025.acl-long.288}
}

@article{li2024geometry,
  title   = {The Geometry of Concepts: Sparse Autoencoder Feature Structure},
  author  = {Li, Yuxiao and Michaud, Eric J. and Baek, David D. and Engels, Joshua and Sun, Xiaoqing and Tegmark, Max},
  journal = {arXiv preprint arXiv:2410.19750},
  year    = {2024},
  url     = {https://arxiv.org/abs/2410.19750}
}

@article{zheng2025mtm,
  title   = {Model Directions, Not Words: Mechanistic Topic Models Using Sparse Autoencoders},
  author  = {Zheng, Carolina and Beltran-Velez, Nicolas and Karlekar, Sweta and Shi, Claudia and Nazaret, Achille and Mallik, Asif and Feder, Amir and Blei, David M.},
  journal = {arXiv preprint arXiv:2507.23220},
  year    = {2025},
  url     = {https://arxiv.org/abs/2507.23220}
}

@inproceedings{marks2024,
  title     = {Sparse Feature Circuits: Discovering and Editing Interpretable Causal Graphs in Language Models},
  author    = {Marks, Samuel and Rager, Can and Michaud, Eric J. and Belinkov, Yonatan and Bau, David and Mueller, Aaron},
  booktitle = {International Conference on Learning Representations},
  year      = {2025},
  url       = {https://openreview.net/forum?id=I4e82CIDxv}
}

@inproceedings{dunefsky2024,
  title     = {Transcoders Find Interpretable LLM Feature Circuits},
  author    = {Dunefsky, Jacob and Chlenski, Philippe and Nanda, Neel},
  booktitle = {Advances in Neural Information Processing Systems},
  volume    = {37},
  year      = {2024},
  url       = {https://proceedings.neurips.cc/paper_files/paper/2024/hash/2b8f4db0464cc5b6e9d5e6bea4b9f308-Abstract-Conference.html}
}

@article{ameisen2025circuit,
  title   = {Circuit Tracing: Revealing Computational Graphs in Language Models},
  author  = {Ameisen, Emmanuel and Lindsey, Jack and Pearce, Adam and Gurnee, Wes and Turner, Nicholas L. and Chen, Brian and Citro, Craig and others},
  journal = {Transformer Circuits Thread},
  year    = {2025},
  url     = {https://transformer-circuits.pub/2025/attribution-graphs/methods.html}
}

@inproceedings{hanna2025circuittracer,
  title     = {Circuit-Tracer: A New Library for Finding Feature Circuits},
  author    = {Hanna, Michael and Piotrowski, Mateusz and Lindsey, Jack and Ameisen, Emmanuel},
  booktitle = {Proceedings of the 8th BlackboxNLP Workshop: Analyzing and Interpreting Neural Networks for NLP},
  pages     = {239--249},
  address   = {Suzhou, China},
  publisher = {Association for Computational Linguistics},
  year      = {2025},
  url       = {https://aclanthology.org/2025.blackboxnlp-1.14/},
  doi       = {10.18653/v1/2025.blackboxnlp-1.14}
}

@article{laptev2025featureflow,
  title   = {Analyze Feature Flow to Enhance Interpretation and Steering in Language Models},
  author  = {Laptev, Daniil and Balagansky, Nikita and Aksenov, Yaroslav and Gavrilov, Daniil},
  journal = {arXiv preprint arXiv:2502.03032},
  year    = {2025},
  url     = {https://arxiv.org/abs/2502.03032}
}

@inproceedings{bollacker2008freebase,
  title     = {Freebase: a collaboratively created graph database for structuring human knowledge},
  author    = {Bollacker, Kurt D. and Evans, Colin and Paritosh, Praveen and Sturge, Tim and Taylor, Jamie},
  booktitle = {Proceedings of the 2008 ACM SIGMOD International Conference on Management of Data},
  pages     = {1247--1250},
  year      = {2008},
  publisher = {ACM},
  doi       = {10.1145/1376616.1376746}
}

@inproceedings{carlson2010nell,
  title     = {Toward an Architecture for Never-Ending Language Learning},
  author    = {Carlson, Andrew and Betteridge, Justin and Kisiel, Bryan and Settles, Burr and Hruschka Jr., Estevam R. and Mitchell, Tom M.},
  booktitle = {Proceedings of the Twenty-Fourth AAAI Conference on Artificial Intelligence},
  pages     = {1306--1313},
  year      = {2010}
}

@article{vrandecic2014wikidata,
  title   = {Wikidata: a free collaborative knowledgebase},
  author  = {Vrande{\v{c}}i{\'c}, Denny and Kr{\"o}tzsch, Markus},
  journal = {Communications of the ACM},
  volume  = {57},
  number  = {10},
  pages   = {78--85},
  year    = {2014},
  doi     = {10.1145/2629489}
}

@inproceedings{mintz2009distant,
  title     = {Distant supervision for relation extraction without labeled data},
  author    = {Mintz, Mike and Bills, Steven and Snow, Rion and Jurafsky, Daniel},
  booktitle = {Proceedings of the Joint Conference of the 47th Annual Meeting of the ACL and the 4th International Joint Conference on Natural Language Processing of the AFNLP},
  pages     = {1003--1011},
  address   = {Suntec, Singapore},
  year      = {2009},
  publisher = {Association for Computational Linguistics},
  url       = {https://aclanthology.org/P09-1113/}
}

@inproceedings{etzioni2011openie,
  title     = {Open Information Extraction: The Second Generation},
  author    = {Etzioni, Oren and Fader, Anthony and Christensen, Janara and Soderland, Stephen and Mausam},
  booktitle = {Proceedings of the Twenty-Second International Joint Conference on Artificial Intelligence},
  pages     = {3--10},
  year      = {2011}
}

@inproceedings{niklaus2018openie,
  title     = {A Survey on Open Information Extraction},
  author    = {Niklaus, Christina and Cetto, Matthias and Freitas, Andr{\'e} and Handschuh, Siegfried},
  booktitle = {Proceedings of the 27th International Conference on Computational Linguistics},
  pages     = {3866--3878},
  address   = {Santa Fe, New Mexico, USA},
  year      = {2018},
  publisher = {Association for Computational Linguistics},
  url       = {https://aclanthology.org/C18-1326/}
}

@article{martinezrodriguez2020iesemweb,
  title   = {Information extraction meets the Semantic Web: A survey},
  author  = {Mart{\'i}nez-Rodr{\'i}guez, Jos{\'e}-L{\'a}zaro and Hogan, Aidan and L{\'o}pez-Ar{\'e}valo, Ivan},
  journal = {Semantic Web},
  volume  = {11},
  number  = {2},
  pages   = {255--335},
  year    = {2020},
  doi     = {10.3233/SW-180333}
}

@article{balcells2024evolution,
  author  = {Balcells, Daniel and Lerner, Benjamin and Oesterle, Michael and Ucar, Ediz and Heimersheim, Stefan},
  title   = {Evolution of SAE Features Across Layers in LLMs},
  journal = {arXiv preprint arXiv:2410.08869},
  year    = {2024},
  doi     = {10.48550/arXiv.2410.08869},
  url     = {https://arxiv.org/abs/2410.08869}
}

@misc{clarke2024saecooccurrence,
  author       = {Clarke, Matthew A. and Bhatnagar, Hardik and Bloom, Joseph},
  title        = {Compositionality and Ambiguity: Latent Co-occurrence and Interpretable Subspaces},
  year         = {2024},
  howpublished = {LessWrong},
  month        = dec,
  url          = {https://www.lesswrong.com/posts/WNoqEivcCSg8gJe5h/compositionality-and-ambiguity-latent-co-occurrence-and}
}

@inproceedings{winnicki2024matrix,
  author    = {John Winnicki and Fr{\'e}d{\'e}ric Poitevin and Haoyuan Li and Eric Darve},
  title     = {Matrix Sketching for Online Analysis of {LCLS} Imaging Datasets},
  booktitle = {{SC24-W:} Workshops of the International Conference for High Performance Computing, Networking, Storage and Analysis},
  pages     = {2144--2153},
  publisher = {{IEEE}},
  year      = {2024},
  doi       = {10.1109/SCW63240.2024.00269},
  url       = {https://doi.org/10.1109/SCW63240.2024.00269}
}

@inproceedings{epstein2025sdkde,
  author    = {Epstein, Elliot L. and Dwaraknath, Rajat Vadiraj and Sornwanee, Thanawat and Winnicki, John and Liu, Jerry Weihong},
  title     = {SD-KDE: Score-Debiased Kernel Density Estimation},
  booktitle = {Advances in Neural Information Processing Systems},
  year      = {2025},
  url       = {https://openreview.net/forum?id=fvho95EtPu}
}

@article{epstein2026flashsdkde,
  title         = {Flash-SD-KDE: Accelerating SD-KDE with Tensor Cores},
  author        = {Epstein, Elliot L. and Dwaraknath, Rajat Vadiraj and Winnicki, John},
  journal       = {arXiv preprint arXiv:2602.10378},
  year          = {2026},
  doi           = {10.48550/arXiv.2602.10378},
  archivePrefix = {arXiv},
  eprint        = {2602.10378},
  primaryClass  = {cs.DC},
  url           = {https://arxiv.org/abs/2602.10378}
}

@misc{epstein2025fermieval,
  author        = {Epstein, Elliot L. and Winnicki, John and Sornwanee, Thanawat and Dwaraknath, Rajat},
  title         = {LLMs are Overconfident: Evaluating Confidence Interval Calibration with FermiEval},
  year          = {2025},
  eprint        = {2510.26995},
  archivePrefix = {arXiv},
  doi           = {10.48550/arXiv.2510.26995},
  url           = {https://arxiv.org/abs/2510.26995}
}

@article{winnicki2025lzwaveforms,
  author  = {Winnicki, John and Arthurs, Maris and Anderson, Tyler and O'Shea, Finn H. and Monzani, Maria Elena and Darve, Eric},
  title   = {Unsupervised Learning Techniques for Identification of Anomalous {LZ} Waveform Data},
  journal = {EPJ Web of Conferences},
  volume  = {337},
  pages   = {01122},
  year    = {2025},
  doi     = {10.1051/epjconf/202533701122},
  url     = {https://doi.org/10.1051/epjconf/202533701122}
}

@article{jolliffe2016pca,
  author  = {Jolliffe, Ian T. and Cadima, Jorge},
  title   = {Principal Component Analysis: A Review and Recent Developments},
  journal = {Philosophical Transactions of the Royal Society A: Mathematical, Physical and Engineering Sciences},
  volume  = {374},
  number  = {2065},
  pages   = {20150202},
  year    = {2016},
  doi     = {10.1098/rsta.2015.0202},
  url     = {https://doi.org/10.1098/rsta.2015.0202}
}

@article{vandermaaten2008tsne,
  author  = {van der Maaten, Laurens and Hinton, Geoffrey},
  title   = {Visualizing Data Using t-SNE},
  journal = {Journal of Machine Learning Research},
  volume  = {9},
  number  = {86},
  pages   = {2579--2605},
  year    = {2008},
  url     = {https://www.jmlr.org/papers/v9/vandermaaten08a.html}
}

@article{mcinnes2018umap,
  author  = {McInnes, Leland and Healy, John and Saul, Nathaniel and Gro{\ss}berger, Lukas},
  title   = {UMAP: Uniform Manifold Approximation and Projection},
  journal = {Journal of Open Source Software},
  volume  = {3},
  number  = {29},
  pages   = {861},
  year    = {2018},
  doi     = {10.21105/joss.00861},
  url     = {https://doi.org/10.21105/joss.00861}
}

@inproceedings{turpin2023unfaithful,
  author    = {Turpin, Miles and Michael, Julian and Perez, Ethan and Bowman, Samuel},
  title     = {Language Models Don't Always Say What They Think: Unfaithful Explanations in Chain-of-Thought Prompting},
  booktitle = {Advances in Neural Information Processing Systems},
  volume    = {36},
  year      = {2023},
  url       = {https://proceedings.neurips.cc/paper_files/paper/2023/hash/ed3fea9033a80fea1376299fa7863f4a-Abstract-Conference.html}
}

@article{lanham2023faithfulness,
  author  = {Lanham, Tamera and Chen, Anna and Radhakrishnan, Ansh and Steiner, Benoit and Denison, Carson and Hernandez, Danny and Li, Dustin and Durmus, Esin and Hubinger, Evan and Kernion, Jackson and others},
  title   = {Measuring Faithfulness in Chain-of-Thought Reasoning},
  journal = {arXiv preprint arXiv:2307.13702},
  year    = {2023},
  doi     = {10.48550/arXiv.2307.13702},
  url     = {https://arxiv.org/abs/2307.13702}
}

@article{wang2023pinto,
  title={PINTO: Faithful Language Reasoning Using Prompt-Generated Rationales},
  author={Wang, Peifeng and Chan, Aaron and Ilievski, Filip and Chen, Muhao and Ren, Xiang},
  journal={arXiv preprint arXiv:2211.01562},
  year={2023}
}

@article{abdollahi2025demystifying,
  title={Demystifying Errors in LLM Reasoning Traces: An Empirical Study of Code Execution Simulation},
  author={Abdollahi, Mohammad and Tasnia, Khandaker Rifah and Saha, Soumit Kanti and Yang, Jinqiu and Wang, Song and Hemmati, Hadi},
  journal={arXiv preprint arXiv:2512.00215},
  year={2025}
}

@misc{lin2023neuronpedia,
  title        = {Neuronpedia: Interactive Reference and Tooling for Analyzing Neural Networks},
  author       = {Lin, Johnny},
  year         = {2023},
  howpublished = {\url{https://www.neuronpedia.org}},
  note         = {Software available from neuronpedia.org}
}

@misc{bills2023language,
  title        = {Language Models Can Explain Neurons in Language Models},
  author       = {Bills, Steven and Cammarata, Nick and Mossing, Dan and Tillman, Henk and Gao, Leo and Goh, Gabriel and Sutskever, Ilya and Leike, Jan and Wu, Jeff and Saunders, William},
  year         = {2023},
  howpublished = {\url{https://openaipublic.blob.core.windows.net/neuron-explainer/paper/index.html}}
}

@book{molnar2015conceptsbiology,
  title     = {Concepts of Biology: 1st Canadian Edition},
  author    = {Molnar, Charles and Gair, Jane},
  year      = {2015},
  publisher = {BCcampus},
  address   = {Victoria, B.C.},
  url       = {https://opentextbc.ca/biology/},
  note      = {Licensed under CC BY 4.0. Adapted from OpenStax, \emph{Concepts of Biology}.}
}

@book{earle2015physicalgeology,
  title     = {Physical Geology},
  author    = {Earle, Steven},
  year      = {2015},
  publisher = {BCcampus},
  address   = {Victoria, B.C.},
  url       = {https://opentextbc.ca/geology/},
  note      = {Licensed under CC BY 4.0.}
}

@book{allosso2021modernworldhistory,
  title  = {Modern World History},
  author = {Allosso, Dan and Williford, Tom},
  year   = {2021},
  url    = {https://mlpp.pressbooks.pub/modernworldhistory/},
  note   = {Licensed under CC BY-NC-SA 4.0.}
}

@article{cunningham2023sparse,
  title   = {Sparse Autoencoders Find Highly Interpretable Features in Language Models},
  author  = {Cunningham, Hoagy and Ewart, Aidan and Riggs, Logan and Huben, Robert and Sharkey, Lee},
  year    = {2023},
  journal = {arXiv preprint arXiv:2309.08600},
  url     = {https://arxiv.org/abs/2309.08600}
}

@article{bricken2023monosemanticity,
  title   = {Towards Monosemanticity: Decomposing Language Models With Dictionary Learning},
  author  = {Bricken, Trenton and Templeton, Adly and Batson, Joshua and Chen, Brian and Jermyn, Adam and Conerly, Tom and Turner, Nick and Anil, Cem and Denison, Carson and Askell, Amanda and Lasenby, Robert and Wu, Yifan and Kravec, Shauna and Schiefer, Nicholas and Maxwell, Tim and Joseph, Nicholas and Hatfield-Dodds, Zac and Tamkin, Alex and Nguyen, Karina and McLean, Brayden and Burke, Josiah E. and Hume, Tristan and Carter, Shan and Henighan, Tom and Olah, Christopher},
  year    = {2023},
  journal = {Transformer Circuits Thread},
  url     = {https://transformer-circuits.pub/2023/monosemantic-features/index.html}
}

@article{templeton2024scaling,
  title   = {Scaling Monosemanticity: Extracting Interpretable Features from Claude 3 Sonnet},
  author  = {Templeton, Adly and Conerly, Tom and Marcus, Jonathan and Lindsey, Jack and Bricken, Trenton and Chen, Brian and Pearce, Adam and Citro, Craig and Ameisen, Emmanuel and Jones, Andy and Cunningham, Hoagy and Turner, Nicholas L. and McDougall, Callum and MacDiarmid, Monte and Tamkin, Alex and Durmus, Esin and Hume, Tristan and Mosconi, Francesco and Freeman, C. Daniel and Sumers, Theodore R. and Rees, Edward and Batson, Joshua and Jermyn, Adam and Carter, Shan and Olah, Christopher and Henighan, Tom},
  year    = {2024},
  journal = {Transformer Circuits Thread},
  url     = {https://transformer-circuits.pub/2024/scaling-monosemanticity/index.html}
}

@misc{karvonen2025saebench,
  title         = {SAEBench: A Comprehensive Benchmark for Sparse Autoencoders in Language Model Interpretability},
  author        = {Karvonen, Adam and Rager, Can and Lin, Johnny and Tigges, Curt and Bloom, Joseph and Chanin, David and Lau, Yeu-Tong and Farrell, Eoin and McDougall, Callum and Ayonrinde, Kola and Wearden, Matthew and Conmy, Arthur and Marks, Samuel and Nanda, Neel},
  year          = {2025},
  eprint        = {2503.09532},
  archivePrefix = {arXiv},
  primaryClass  = {cs.LG},
  url           = {https://arxiv.org/abs/2503.09532}
}

\appendix

\section{Additional Figures}
\label{app:extra}

\paragraph{Interactive artifact.}
An interactive browser for the dual-graph artifact is available at \url{https://saegraph-render-dual-private.onrender.com/}. 

\begin{figure}[H]
    \centering
    \includegraphics[width=\textwidth]{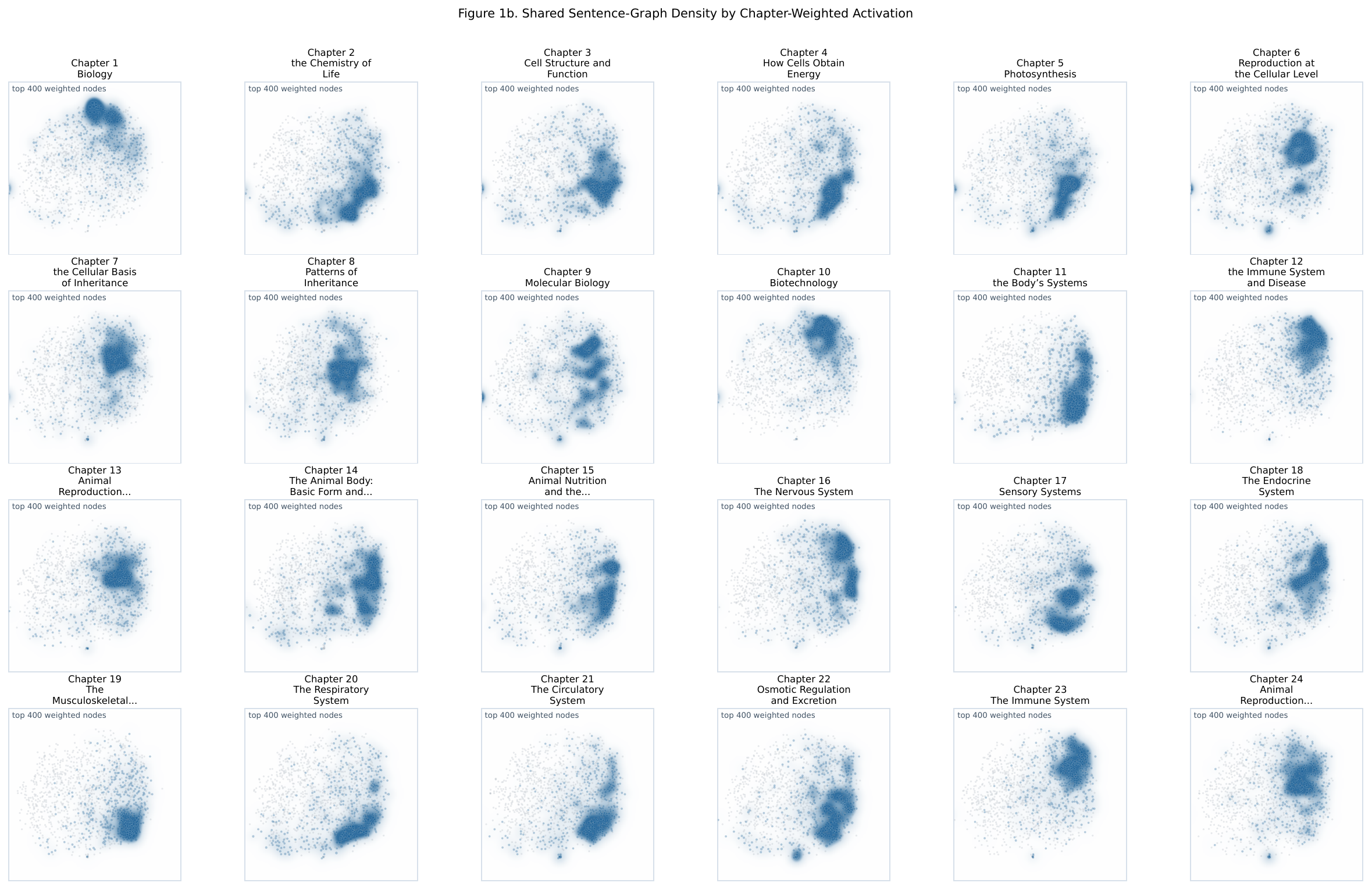}
    \caption{Full chapter-density atlas on shared sentence-graph coordinates. Each panel shows one chapter's weighted view inside the same strict sentence graph.}
    \label{fig:chapter-density-full}
\end{figure}

\section{Filter Implementation Details}
\label{app:filter-details}

\paragraph{Shortlist mechanics.}
The shortlist first applies a target-activity gate so that vanishingly weak features never reach packet review. In the biology run used here, the config requires minimum target support rate $5\times 10^{-4}$, minimum target activation mass $10$, and removal of the bottom $20\%$ of target-support features within each layer before ranking. Surviving features are then ranked by a recall-oriented combined score over target-versus-contrast enrichment, within-book localization, and a small synergy term. The current config uses subchapter-level binary-support localization with a strong-support hybrid refinement, and exports the top 30k features as the shortlist.

\paragraph{What enters a packet.}
With the shortlist in hand, we package each shortlisted feature as an auditable evidence record. Each packet includes the external feature description, a local description-screen label (\texttt{candidate}, \texttt{surface\_form}, or \texttt{missing}), target-side statistics and evidence sentences, contrast-side statistics, activated and comparator chapter/subchapter units, chapter ids and titles, shortlist reason tags, and optional nearby-duplicate hints. Missing and surface-form features are filtered locally before any API call.

\paragraph{Current biology adjudication contract.}
For the default biology path, adjudication uses a run-level domain profile together with the adjudication contract. The API returns two schema-validated outputs per feature: a visibility record containing \texttt{visible} and \texttt{evidence\_sentence\_ids}, and a relevance record containing \texttt{belongs\_here}, \texttt{distinctiveness}, and a one-sentence justification. These correspond to the three adjudication questions: is the concept visible in the evidence sentences, does it belong in the activated biology region, and is that fit more distinctive than the comparator fit?

\section{Detailed hierarchy and mechanism methods}
\label{app:methods-detail}

\subsection{The abstraction tree over the retained concept universe}
\label{sec:abstraction-tree}

Let $\mathcal{V}^{\star}=\{1,\dots,F^{\star}\}$ denote the retained readable concept universe, and write $\mathcal{V}:=\mathcal{V}^{\star}$ for notational convenience below. Each feature $v\in\mathcal{V}$ has a text description $d_v$. Rather than treat that vocabulary as a flat list, we organize it into an abstraction tree whose leaves are individual features and whose internal nodes are grounded semantic groupings. The tree is useful for two reasons. First, it gives a researcher a manageable way to navigate the part of feature space that survived the strict filter. Second, once a grouped view is chosen inside that retained universe, the same leaf subset can be reused in every later view of the paper.

\subsubsection{Building a grounded abstraction tree}

Each retained feature $v\in\mathcal{V}$ has a text description $d_v$. We embed those descriptions, project them into a low-dimensional PCA space \citep{jolliffe2016pca}, and build a $k$-nearest-neighbor graph in that space. We then keep only mutual-$k$NN edges. This removes many one-sided hub connections and gives a cleaner local neighborhood structure for the recursive splitting step.

The tree is built recursively from that geometry. At a current node, we choose a branching factor from the node size, place seeds by farthest-first coverage, and assign features to the nearest seed while preferring assignments that stay connected in the mutual-$k$NN graph. If a candidate child is too diffuse or too weakly connected, we reseed and regrow it. This makes the internal nodes more coherent than what we would get from a single flat clustering pass.

Clustering alone is not enough. After each split, we summarize the parent from grounded evidence: representative children near the centroid, semantic extremes that widen coverage, near-boundary negatives that sharpen scope, and descendant leaf anchors that keep the summary tied to concrete features. The summarization step is asked to produce a concept that subsumes its children and is more general than any one child, without drifting away from the leaves. This is what makes the hierarchy useful as an abstraction tree rather than just a partition of the feature cloud.

\begin{figure*}[t]
    \centering
    \includegraphics[width=\textwidth]{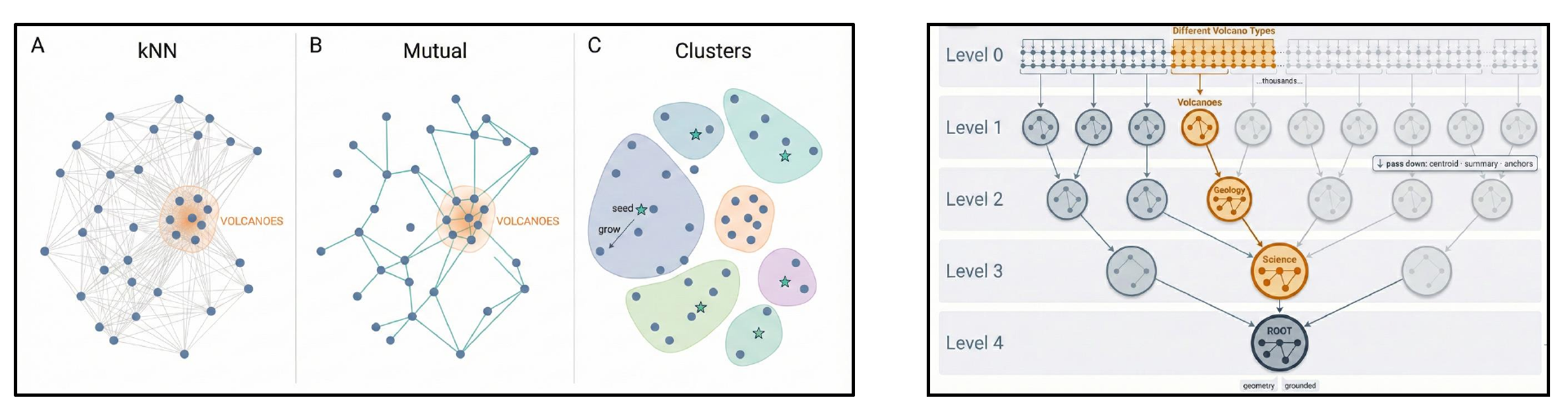}
    \caption{Constructing the abstraction tree. \textbf{(a)} A mutual-$k$NN filter cleans the local neighborhood graph built from feature-description embeddings, which gives a better geometry for recursive splitting. \textbf{(b)} The resulting local groups are organized into progressively broader internal concepts, with summaries grounded by representatives and descendant leaf anchors.}
    \label{fig:abstraction-tree}
\end{figure*}

Figure~\ref{fig:abstraction-tree} shows the construction. Panel~\textbf{(a)} shows the local neighborhood geometry after the mutual-$k$NN filter, which is the object we actually split. Panel~\textbf{(b)} shows the recursive hierarchy that results from repeatedly grouping local neighborhoods into broader internal concepts.

\subsubsection{Optional hierarchy-defined grouped subviews and downstream reuse}

Although the strict filter now fixes the retained universe, the hierarchy still supports reusable grouped views inside that universe. If $U \subseteq \mathcal{I}(\mathcal{T})$ is a selected set of internal nodes, we export
\[
\slice{U} = \bigcup_{u\in U} \descop(u) \subseteq \mathcal{V}.
\]

This is usually more useful than choosing a single internal node. In practice, someone studying a model behavior may want to keep several concept families at once while dropping retained regions that are still irrelevant to the question at hand. The abstraction tree makes that grouped-view choice fast, explicit, and reusable.

This is also where the hierarchy acts as semantic compression. The original feature space may be enormous, but a small set of named internal nodes can define a compact grouped view that still preserves the leaf features needed for later analysis. Because that grouped view is tied to internal nodes rather than to a one-off hand filter, it can be audited, reused across examples, and compared across views.

\subsection{Multi-granular co-occurrence graphs inside the retained universe}

Let $\mathcal{V}_{\mathrm{work}} \subseteq \mathcal{V}^{\star}$ denote the node universe currently under study. In the default case, $\mathcal{V}_{\mathrm{work}}=\mathcal{V}^{\star}$; when a hierarchy-defined grouped view is requested, $\mathcal{V}_{\mathrm{work}}=\slice{U}$.

\subsubsection{From the working universe to sentence presence}

Once the retained universe (or an optional grouped view inside it) has been fixed, the next question is statistical: which concepts inside that region tend to recur together in the corpus? The co-occurrence view answers that question. It is built on the same selected node universe used throughout the rest of the paper, so we do not first build a full-vocabulary graph and only later trim it for display. The working universe is fixed first, and the co-occurrence graph is constructed inside that region.

The first step is to decide when a feature in $\mathcal{V}_{\mathrm{work}}$ is genuinely present in a sentence. Our pipeline uses a presence-first rule. Let $h_{i,v}$ denote the token-level activation of feature $v$ on token $i$, and let $h_{i,v}^{+}=\max(h_{i,v},0)$. For sentence $s$, we score feature $v$ by its strongest positive token activation in that sentence, after excluding special tokens:
\[
 m_{s,v} = \max_{i\in s} h_{i,v}^{+}.
\]
A sentence feature is then marked present when this score exceeds a feature-specific threshold,
\[
X^{(\mathrm{sent})}_{s,v}(\mathcal{V}_{\mathrm{work}})
=
\mathbf{1}\!\left[m_{s,v} > \theta_v\right],
\qquad v\in \mathcal{V}_{\mathrm{work}},
\]
where $\theta_v$ is calibrated from the empirical distribution of that feature's sentence scores, with rare-feature safeguards.

This detail matters for interpretation. Presence is not decided by ranking features against one another within a sentence. A feature can enter the sentence state because it fires strongly on one token relative to its own corpus behavior. Co-occurrence therefore begins from feature-wise presence events, not from a sentence-level competition among features.

\subsubsection{Lifting presence to larger units and building the graph}

Sentence presence is the primitive event. Larger textual units are built deterministically from it. For granularity $g\in\{\mathrm{paragraph},\mathrm{subchapter},\mathrm{chapter}\}$ and unit $u$ with constituent sentences $\mathcal{C}_g(u)$, we define
\[
X^{(g)}_{u,v}(\mathcal{V}_{\mathrm{work}})
=
\mathbf{1}\!\left[\exists s\in \mathcal{C}_g(u) : X^{(\mathrm{sent})}_{s,v}(\mathcal{V}_{\mathrm{work}})=1\right].
\]
This gives a family of binary presence matrices $X^{(g)}(\mathcal{V}_{\mathrm{work}})$ whose columns are exactly the selected features in $\mathcal{V}_{\mathrm{work}}$. The different granularities therefore answer the same question at different scales instead of redefining the feature set each time.

At any chosen granularity, co-occurrence counts are computed from the restricted presence matrix,
\[
C^{(g)}(\mathcal{V}_{\mathrm{work}}) = X^{(g)}(\mathcal{V}_{\mathrm{work}})^{\top} X^{(g)}(\mathcal{V}_{\mathrm{work}}).
\]
We then use Jaccard normalization to discount very common features,
\[
J^{(g)}_{ab}(\mathcal{V}_{\mathrm{work}})
=
\frac{C^{(g)}_{ab}(\mathcal{V}_{\mathrm{work}})}{C^{(g)}_{aa}(\mathcal{V}_{\mathrm{work}})+C^{(g)}_{bb}(\mathcal{V}_{\mathrm{work}})-C^{(g)}_{ab}(\mathcal{V}_{\mathrm{work}})},
\]
and extract a sparse same-layer view graph by keeping strong neighbors per node and symmetrizing. The result is the co-occurrence graph $G^{\cooc}_g(\mathcal{V}_{\mathrm{work}})$.

Two points are important here. First, after sentence presence is defined, the construction is purely combinatorial: one stage counts pairwise co-presence and another sparsifies the weighted matrix. Second, sparsification is performed after restricting to $\mathcal{V}_{\mathrm{work}}$. The retained edges are therefore the strongest associations inside the selected semantic region, not merely the edges that survive from a larger global graph.

\subsubsection{What this view contributes}

The filter fixes where to look, and the hierarchy can optionally define a grouped subview inside that universe. The co-occurrence graph answers what tends to appear together once that region has been chosen. Some edges reveal stable bundles of concepts that repeatedly recur together. Some nodes act as bridges between neighborhoods. Small motifs such as wedges and triangles show whether the model tends to deploy isolated pairings, hub-like concept packages, or tightly coupled groups.

This view is useful because the strict filter has already removed much of the irrelevant clutter. Inside a good retained universe (or inside a good grouped subview of it) same-layer co-occurrence provides a corpus-level map of recurring conceptual organization. Later sections add evidence-backed relation labels to those edges and compare this statistical map with the directed mechanism view.

\subsection{Transcoder mechanism knowledge graphs}
\label{sec:transcoder}

\subsubsection{Transcoders as sparse residual-to-residual mechanisms}

The co-occurrence graph tells us which readable concepts statistically belong together inside a selected node universe. The transcoder view is introduced for a different question: on a particular input, which source-side concepts appear to lead to which target-side concepts across a layer transition? The standard transcoder object is a sparse residual-to-residual mechanism. For token $i$, let
\[
 x_i^{\mathrm{src}} \in \mathbb{R}^{d}
 \qquad\text{and}\qquad
 x_i^{\mathrm{tgt}} \in \mathbb{R}^{d}
\]
denote the source-site and target-site residual vectors. The transcoder computes a sparse latent code
\[
 t_i = \operatorname{ReLU}(R x_i^{\mathrm{src}}),
 \qquad t_i \in \mathbb{R}^{K},
\]
and writes the residual update
\[
 \widehat{\Delta x}_i = W t_i,
 \qquad \widehat{\Delta x}_i \in \mathbb{R}^{d},
\]
where $R \in \mathbb{R}^{K \times d}$ is the read matrix and $W \in \mathbb{R}^{d \times K}$ is the write matrix. Each latent $k$ therefore has a read vector $r_k$ (row $k$ of $R$) and a write vector $w_k$ (column $k$ of $W$). In the usual mechanistic reading, latent $k$ detects one residual pattern and writes another: it is a candidate mechanism that maps part of the source residual stream into a target-side residual update.

That residual-level picture is the right starting point, but it is not yet readable enough for scientific inspection. The goal of this section is to place human-interpretable SAE dictionaries on both sides of the transcoder so that the same residual-to-residual mechanism can be viewed as a mapping from readable source-side features to readable target-side features. Concretely,
\[
\underbrace{x_i^{\mathrm{src}}}_{\text{source residual}}
\longrightarrow
\underbrace{t_i}_{\text{transcoder mechanism}}
\longrightarrow
\underbrace{\widehat{\Delta x}_i}_{\text{target residual update}}
\]
is turned into
\[
\underbrace{z_i^{\mathrm{src}}}_{\text{readable SAE source features}}
\longrightarrow
\underbrace{t_i}_{\text{mechanism latents}}
\longrightarrow
\underbrace{z_i^{\mathrm{tgt}}}_{\text{readable SAE target features}}.
\]
The paper therefore treats each latent as the mechanism object, while readable feature-to-feature edges are derived views used for retrieval, aggregation, and browsing.

\begin{figure}[t]
    \centering
    \includegraphics[width=\linewidth]{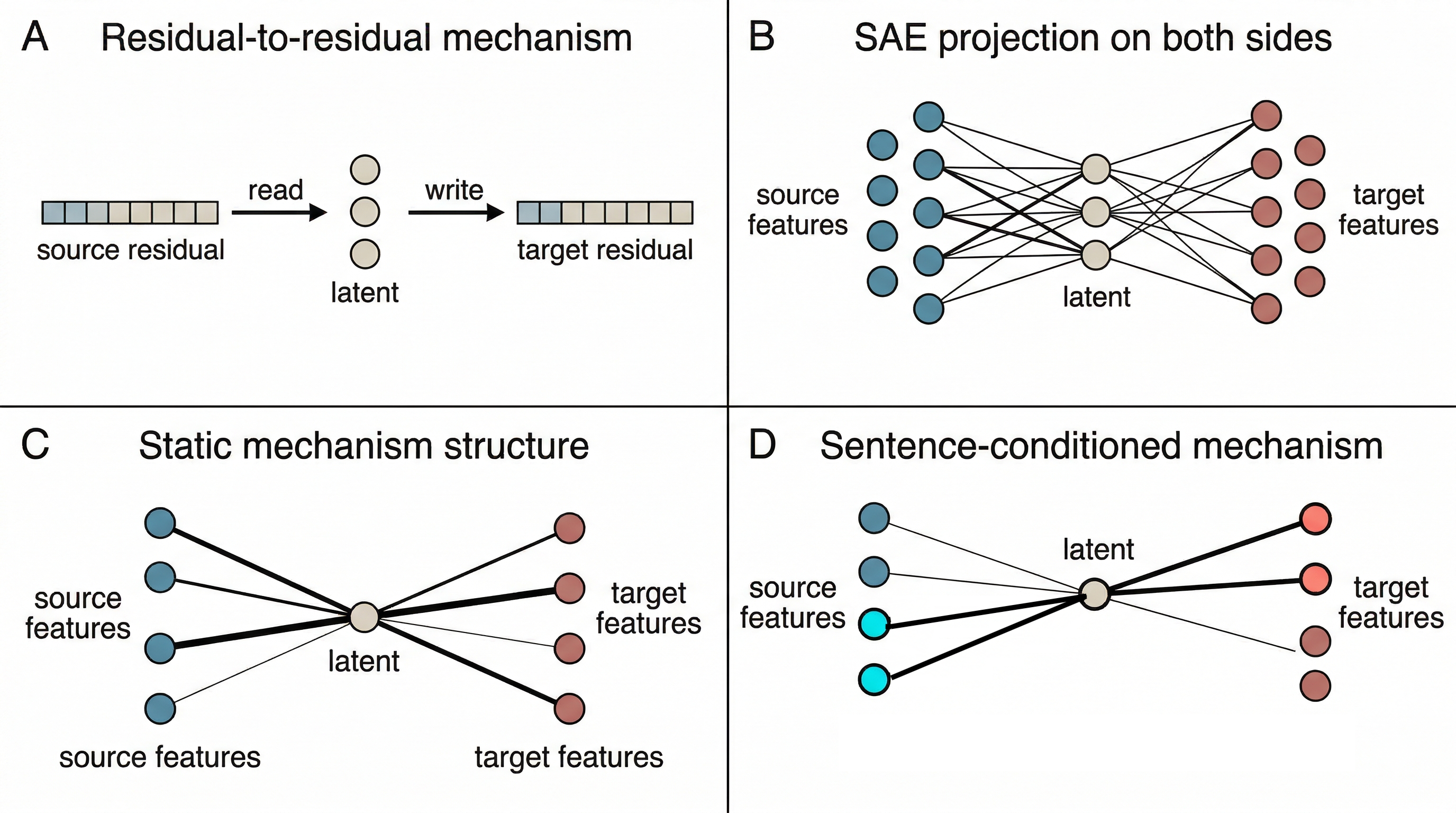}
    \caption{
    Transcoder mechanisms viewed through readable SAE dictionaries.
    (\textbf{A}) A standard transcoder maps source residual patterns to target residual updates through sparse latents.
    (\textbf{B}) Placing SAE dictionaries on both sides turns that residual-space mechanism into a mapping between readable source-side and target-side features.
    (\textbf{C}) Each latent induces a static mechanism profile: source features it tends to read from and target features it tends to write toward.
    (\textbf{D}) On a selected sentence, only a subset of those source features, latents, and target features are active, yielding the sentence-conditioned mechanism view used for local explanation.
    }
    \label{fig:transcoder_overview}
\end{figure}

Figure~\ref{fig:transcoder_overview} gives the high-level picture for the transcoder view. The starting point is the standard residual-to-residual mechanism in Panel~A: a sparse latent reads a source-side residual pattern and writes a target-side residual update. The key representational step of this paper is shown in Panel~B, where SAE dictionaries are placed on both sides so that the same latent can be inspected as a mapping between readable source-side and target-side features. Panels~C and~D then separate the two objects used below: a static mechanism profile, which says what a latent is positioned to read and write, and a dynamic sentence-conditioned mechanism, which says which of those candidate connections are actually active on the selected input.

\subsubsection{Residual-space objects and readable feature dictionaries}

We use one SAE dictionary on the source side and one on the target side. Let
\[
E^{\mathrm{src}} \in \mathbb{R}^{F_{\mathrm{src}} \times d},
\qquad
D^{\mathrm{src}} \in \mathbb{R}^{d \times F_{\mathrm{src}}},
\]
\[
E^{\mathrm{tgt}} \in \mathbb{R}^{F_{\mathrm{tgt}} \times d},
\qquad
D^{\mathrm{tgt}} \in \mathbb{R}^{d \times F_{\mathrm{tgt}}}
\]
denote the source-side and target-side SAE encoders and decoders. The readable sparse activations are
\[
 z_i^{\mathrm{src}} = \operatorname{ReLU}(E^{\mathrm{src}} x_i^{\mathrm{src}}),
 \qquad
 z_i^{\mathrm{tgt}} = \operatorname{ReLU}(E^{\mathrm{tgt}} x_i^{\mathrm{tgt}}).
\]
Write
\[
 d_a^{\mathrm{src}} := D^{\mathrm{src}}_{:,a}
 \qquad\text{and}\qquad
 e_b^{\mathrm{tgt}} := E^{\mathrm{tgt}}_{b,:}
\]
for the source-feature decoder direction of feature $a$ and the target-feature encoder direction of feature $b$. These are the readable dictionary objects that we use to interpret the transcoder's read and write vectors.

The remainder of the section separates three objects that should not be conflated:
\begin{enumerate}[leftmargin=1.5em,itemsep=0.15em,topsep=0.2em]
    \item a \emph{static mechanism} for each latent, which says what that latent is positioned to read from and write toward in the SAE dictionaries;
    \item a \emph{dynamic mechanism execution} on a selected analysis unit, which says which of those candidate mechanisms were actually used on that input;
    \item a \emph{compressed dynamic graph}, which uses the semantic hierarchy to keep dense local mechanism graphs readable without hiding stored active edges.
\end{enumerate}

\subsubsection{Functional source and target support are local linear proxies}

The source-side question is: which readable source features tend to support the firing of latent $k$? The latent preactivation is
\[
 p_{i,k} = r_k^{\top} x_i^{\mathrm{src}}.
\]
Using the source-side SAE reconstruction,
\[
 x_i^{\mathrm{src}} \approx D^{\mathrm{src}} z_i^{\mathrm{src}} = \sum_{a=1}^{F_{\mathrm{src}}} z_{i,a}^{\mathrm{src}} d_a^{\mathrm{src}},
\]
we obtain the local linear decomposition
\[
 p_{i,k}
 \approx
 \sum_{a=1}^{F_{\mathrm{src}}} z_{i,a}^{\mathrm{src}} \, \langle d_a^{\mathrm{src}}, r_k \rangle.
\]
This motivates the source-support matrix
\[
 A^{\mathrm{func}}_{a,k} := \langle d_a^{\mathrm{src}}, r_k \rangle,
 \qquad
 A^{\mathrm{func}} \in \mathbb{R}^{F_{\mathrm{src}} \times K}.
\]
Large positive $A^{\mathrm{func}}_{a,k}$ means that source feature $a$ points in a residual direction that tends to increase the preactivation of latent $k$.

The target-side question is: once latent $k$ fires, which readable target features does it tend to support? Since
\[
 \widehat{\Delta x}_i = \sum_{k=1}^{K} t_{i,k} w_k,
\]
the contribution of latent $k$ to the preactivation of target feature $b$ is
\[
 \Delta p_{i,b}^{\mathrm{tgt}}(k)
 :=
 e_b^{\mathrm{tgt}\top}(t_{i,k} w_k)
 =
 t_{i,k} \, \langle e_b^{\mathrm{tgt}}, w_k \rangle.
\]
This motivates the target-support matrix
\[
 G^{\mathrm{func}}_{b,k} := \langle e_b^{\mathrm{tgt}}, w_k \rangle,
 \qquad
 G^{\mathrm{func}} \in \mathbb{R}^{F_{\mathrm{tgt}} \times K}.
\]
Large positive $G^{\mathrm{func}}_{b,k}$ means that firing latent $k$ tends to push the target residual in a direction that raises target feature $b$.

For the graph objects in this paper we use the positive parts
\[
 (A^{\mathrm{func}})^+_{a,k} := \max(A^{\mathrm{func}}_{a,k},0),
 \qquad
 (G^{\mathrm{func}})^+_{b,k} := \max(G^{\mathrm{func}}_{b,k},0),
\]
so the displayed support is explicitly nonnegative and excitatory. These support scores are the linearized interface between the transcoder and the readable SAE dictionaries: they quantify read-side compatibility and write-side compatibility in the shared residual space. They are not intended as a complete runtime decomposition of the full nonlinear computation. Their purpose is narrower and more practical: they are the quantities that make a residual-space mechanism legible as a feature-space mechanism.

\subsubsection{Static mechanisms: dense profiles, readable captions, and the static feature prior}

For each latent $k$, the source-support profile $(A^{\mathrm{func}})^+_{:,k}$ and target-support profile $(G^{\mathrm{func}})^+_{:,k}$ define a dense static mechanism profile. This dense profile is the primary static object. For browsing, storage, and edge explanations, we attach a concise readable caption of the form
\[
 S_k \xrightarrow[\ell_k]{k} T_k,
\]
where $S_k$ is a readable source-side summary, $T_k$ is a readable target-side summary, and $\ell_k$ is a short latent label. The caption is a summary of the dense profile, not a second notion of mechanism.

We use two captioning modes. A simple \emph{top-functional} caption takes the highest positive source and target supports under $(A^{\mathrm{func}})^+_{:,k}$ and $(G^{\mathrm{func}})^+_{:,k}$ directly. A more compact \emph{constrained sparse nonnegative} caption instead fits a short nonnegative description anchored to the same high-scoring functional candidates. In that mode we solve sparse nonnegative approximations such as
\[
 r_k \approx D^{\mathrm{src}} \alpha_k,
 \qquad \alpha_k \ge 0,
\]
\[
 w_k \approx D^{\mathrm{tgt}} \beta_k,
 \qquad \beta_k \ge 0,
\]
with the support of $\alpha_k$ and $\beta_k$ restricted to the strongest functional candidates for latent $k$. Restricting the fit in this way keeps the readable caption anchored to the same features that drive the dynamic evidence equations below.

The static feature prior induced by the full latent collection is then
\[
 M^{\mathrm{static}}_{a,b}
 :=
 \sum_{k=1}^{K} (A^{\mathrm{func}})^+_{a,k} \, (G^{\mathrm{func}})^+_{b,k}.
\]
This prior is useful because it is prompt-independent. It says which source$\to$target feature pairs the transcoder is generically positioned to mediate before any sentence or other analysis unit is selected. The static mechanism library and the static feature prior are therefore complementary static objects: the former is latent-centered, the latter feature-centered.

\subsubsection{Dynamic mechanisms on a selected analysis unit}

Static support tells us what a latent \emph{can} connect. The dynamic mechanism question is different: which of those candidate mechanisms are actually relevant on the selected analysis unit $q$? Let $I_q$ denote the token positions in $q$. We introduce binary presence gates
\[
 g_{i,a}^{\mathrm{src}} \in \{0,1\}
 \qquad\text{and}\qquad
 g_{i,b}^{\mathrm{tgt}} \in \{0,1\},
\]
indicating whether source feature $a$ and target feature $b$ are active on token $i$. We then define token-level latent-specific evidence by multiplying together exactly the terms we want to be simultaneously true:
\begin{align}
E^{\mathrm{lat}}_{i,a,b,k}
:={}&
\underbrace{g_{i,a}^{\mathrm{src}}}_{\text{source feature present}}
\;\underbrace{(A^{\mathrm{func}})^+_{a,k}}_{\text{source aligns with latent read}}
\\
&\times\;
\underbrace{t_{i,k}}_{\text{latent fires on token }i}
\;\underbrace{(G^{\mathrm{func}})^+_{b,k}}_{\text{latent supports target write}}
\\
&\times\;
\underbrace{g_{i,b}^{\mathrm{tgt}}}_{\text{target feature present}}.
\end{align}
This equation is designed so that evidence is high only when the readable source feature is active, the latent is both geometrically relevant and actually firing, and the readable target feature is active on the same token. Aggregating over the selected unit gives
\[
 E^{(q)}_{a,b,k} := \sum_{i \in I_q} E^{\mathrm{lat}}_{i,a,b,k}.
\]
This is the dynamic mechanism score for latent $k$ mediating visible edge $a \to b$ on unit $q$.

Summing over latents yields the displayed edge weight
\[
 F^{(q)}_{a,b} := \sum_{k=1}^{K} E^{(q)}_{a,b,k},
\]
and normalizing over latents yields the explanation distribution
\[
 \rho^{(q)}_{a,b,k}
 :=
 \frac{E^{(q)}_{a,b,k}}{\sum_{j=1}^{K} E^{(q)}_{a,b,j} + \varepsilon}.
\]
 The static objects $A^{\mathrm{func}}$, $G^{\mathrm{func}}$, $S_k \xrightarrow[\ell_k]{k} T_k$, and $M^{\mathrm{static}}$ describe the prompt-independent mechanism library. The dynamic objects $E^{(q)}_{a,b,k}$, $F^{(q)}_{a,b}$, and $\rho^{(q)}_{a,b,k}$ describe which of those mechanisms appear to execute on the chosen unit. The same equations support sentence graphs and larger-unit graphs; only the choice of $q$ changes.

\subsubsection{Hierarchy-respecting compression of dense mechanism graphs}

\begin{figure}[t]
    \centering
    \includegraphics[width=\linewidth]{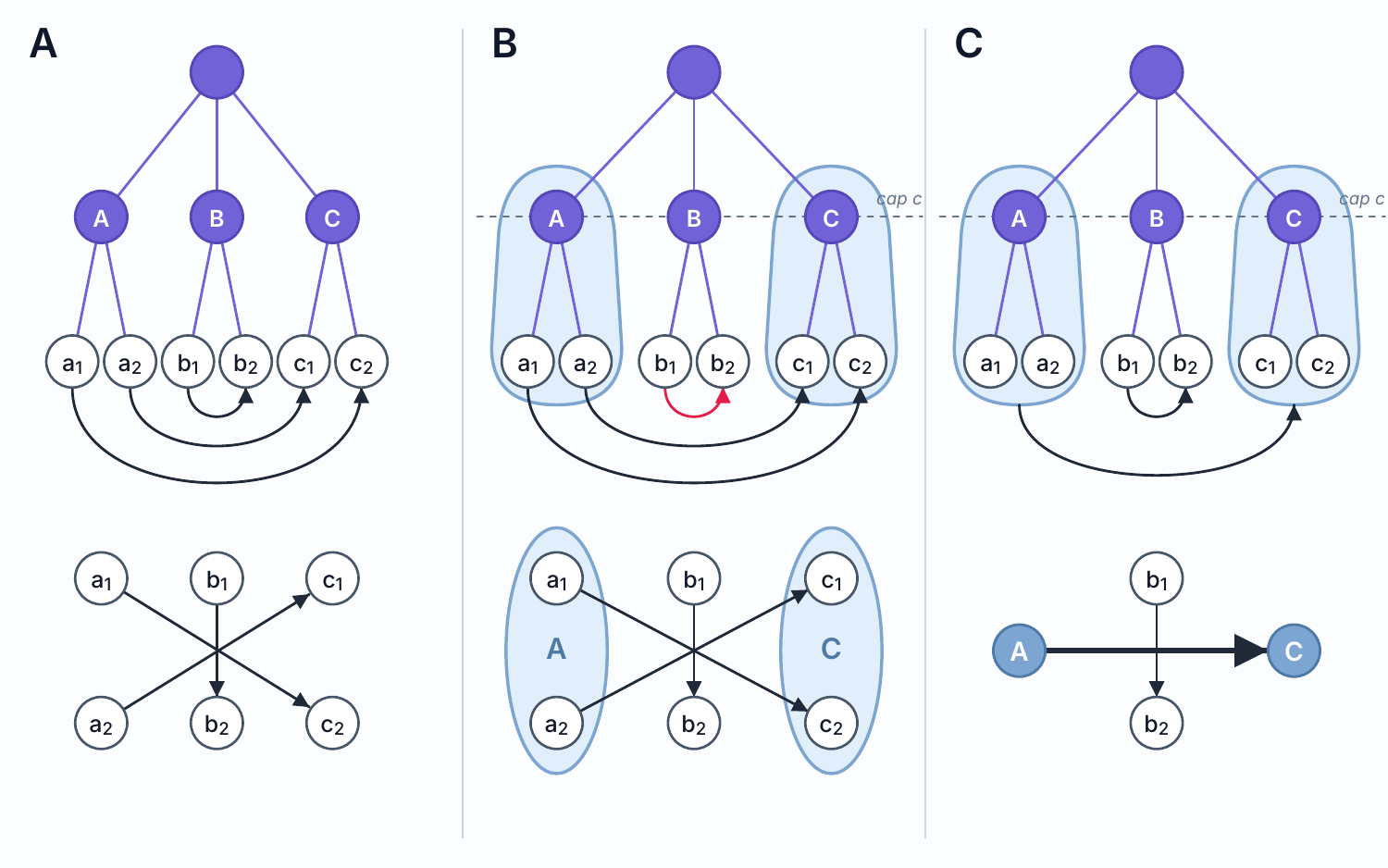}
    \caption{
    Hierarchy-respecting compression of a dense mechanism graph.
    (\textbf{A}) The original view contains leaf-level hierarchy nodes together with leaf-level directed mechanism edges.
    (\textbf{B}) The collapse cap and blocking rule are applied: candidate hierarchy nodes such as $A$ and $C$ are eligible to collapse, while regions crossed by active internal edges remain blocked.
    (\textbf{C}) The displayed graph is formed from a mixed cover of supernodes and leaves. Directed leaf-to-leaf edges between descendants of $A$ and $C$ are aggregated into one displayed superedge, while the internal $b_1 \to b_2$ edge remains explicit.}
    \label{fig:transcoder_compression}
\end{figure}

Even inside a narrow semantic region, the sentence-conditioned mechanism graph can be too dense to read comfortably. We therefore compress the dynamic graph using the same hierarchy that was built over the retained universe. Compression is performed over the shared source+target leaf universe. The dynamic unit payload determines the active leaf set and the blocking edges; the fixed hierarchy determines which regions can collapse.

Concretely, every directed edge stored in the dynamic payload for unit $q$ is treated as a blocking edge. For each stored edge $(u,v)$, we mark the lowest common ancestor of $u$ and $v$ and all of its ancestors as blocked. A hierarchy node may collapse into one displayed supernode only if it (i) contains at least two active descendant leaves, (ii) is not blocked, and (iii) lies below the chosen collapse cap. The resulting mixed cover is then used to project and aggregate leaf-level edges into a smaller displayed graph. 

Figure~\ref{fig:transcoder_compression} illustrates the compression rule used for dense sentence-conditioned mechanism graphs. We begin with leaf-level directed edges over the active subset, mark every stored edge as a blocking edge, and prevent collapse through the lowest common ancestor of each active connection and its ancestors. The remaining eligible hierarchy nodes define a mixed cover of supernodes and leaves, after which leaf-level edges are projected and aggregated into displayed directed edges. As a result, broad cross-group traffic such as $A \to C$ can be compressed into one readable superedge, while internally important structure such as $b_1 \to b_2$ is preserved explicitly.

This matters because it ties the mechanism view back to the hierarchy: the same abstraction structure that helps organize the retained universe also provides the principled compression used to browse its local mechanism structure.

\subsubsection{Interpretation and scope}

The transcoder section defines an explanatory knowledge graph over candidate concept flow. A visible edge $a \to b$ says that, on the selected analysis unit, one or more latents with readable captions jointly support the transition from source-side feature $a$ to target-side feature $b$. The static support matrices make that claim readable in adjacent SAE dictionaries, and the dynamic evidence makes it input-conditioned. This is stronger than a generic feature-correlation story, but it is still not a literal claim of unique causal structure in the base model. The intended interpretation is therefore a useful middle ground: a sentence-grounded, latent-mediated explanation of how one readable concept appears to lead to another.

\section{Edge semantics and Auto-Relate}
\label{app:workflow-detail}

\subsection{Auto-related edge semantics}

\subsubsection{Why weighted edges are not enough}

A weighted edge is not yet a readable scientific claim. A co-occurrence edge tells us that two concepts recur together within a selected node universe, but not what relation is being expressed. A directed mechanism edge tells us that a source-side concept family and a target-side concept family are connected by latent-mediated evidence on a selected input, but not what readable transformation that edge should suggest to a researcher. To turn the two graph families into a usable knowledge graph, we need edge semantics backed by evidence.

\subsubsection{Co-occurrence auto-relate}

For retained co-occurrence edges we build an evidence packet
\[
\mathcal{P}^{(g)}_{ab}
=
\bigl(
\mathcal{J}_{ab},
\mathcal{A}_{a\setminus b},
\mathcal{A}_{b\setminus a},
C^{(g)}_{ab},
J^{(g)}_{ab}
\bigr),
\]
where $\mathcal{J}_{ab}$ stores JOINT evidence lines in which both endpoint features are present, while $\mathcal{A}_{a\setminus b}$ and $\mathcal{A}_{b\setminus a}$ store contrastive lines in which only one endpoint is present.

Labeling proceeds conservatively in two stages. A presence pass checks whether the two endpoint concepts are genuinely supported by the JOINT evidence. A second pass proposes a short connector phrase using JOINT evidence as primary support and the contrastive lines only for disambiguation. Validation rejects phrases that are generic, directionally misleading, or merely copies of the endpoint descriptions, falling back to a conservative relation when the evidence is weak.

To understand why this matters, consider the following real example in labeling a co-occurrence edge between SAE features ``yellow'' and ``national parks.'' A node description may say \emph{yellow}, but the JOINT evidence for the edge to \emph{national park} is dominated by \emph{Yellowstone} examples. Contrastive evidence is exactly what tells the system that the edge is about a place-name usage rather than generic color semantics. In practice, this makes the co-occurrence relation layer evidence-conditioned rather than description-only.

\subsubsection{Directed transcoder auto-relate}

Directed transcoder auto-relate follows the same basic pattern as co-occurrence auto-relate, but the evidence is different because the edge itself is different. A co-occurrence edge is supported by sentences in which two same-layer features fire together. A directed transcoder edge is supported by sentences in which the model shows strong evidence for the specific source$\to$target write.

For a visible edge $a\to b$, we therefore do not rank candidate evidence sentences by endpoint firing rates alone. Instead, we rank them by latent-specific edge evidence from Appendix~\ref{app:methods-detail}. If $E_{a,b,k}^{(q)}$ is the evidence that latent $k$ helped mediate the write from $a$ to $b$ on sentence $q$, then the JOINT evidence for auto-relate is drawn from the sentences with the largest values of $E_{a,b,k}^{(q)}$ over the strongest supporting latents. This keeps the evidence tied to the mechanism edge we are trying to interpret, rather than to broad source-side or target-side activity that may be present for other reasons.

As in the co-occurrence case, we also collect contrastive evidence. Source-only sentences help show what feature $a$ means when it does not strongly write into $b$, and target-only sentences help show what feature $b$ looks like when it is reached through other routes. The evidence packet also includes the label of the strongest supporting transcoder latent. That label is useful as a mechanism hint, but it is not treated as ground truth. It helps the labeling step disambiguate the intended transformation, while the sentence evidence remains primary.

The labeling step is then almost the same as before: first verify that the JOINT evidence really supports the source and target concepts, then propose a short readable relation. The main difference is that the relation must be directional. It should read as a source-to-target transformation or support claim, not as a symmetric topic phrase. Validation should also reject labels that merely copy the endpoint descriptions or the transcoder latent label. The goal is to produce an evidence-backed statement of how one concept family tends to write into another.

\subsubsection{Auto-relate turns the views into knowledge graphs}

The two auto-relate workflows give the two graph families distinct but complementary edge semantics. In the co-occurrence view, labeled edges answer ``what concepts tend to appear together here, and how are they related?'' In the mechanism view, labeled edges answer ``on this selected input, how did one concept family help write into another?'' The result is a typed knowledge graph whose same-layer relations and cross-layer transformations live on the same selected semantic subset.

% \begin{figure}[t]
% \centering
% \fbox{\parbox[c][40mm][c]{0.92\linewidth}{\centering
% \textbf{Figure placeholder}\\[2pt]
% Panel A: co-occurrence edge between a feature described as ``yellow'' and a feature described as ``national park'', with JOINT evidence dominated by Yellowstone examples and A-only/B-only evidence clarifying the intended relation.\\[6pt]
% Panel B: directed transcoder edge from ``eruption / ash plume'' to ``park closure'' on one selected sentence, with JOINT evidence lines, contrastive lines, top latent hint, and the final directed relation label. This figure should make the distinction between same-layer relation labeling and directed transformation labeling visually obvious.}}
% \caption{Auto-relate supplies the edge semantics that turn weighted graphs into readable knowledge graphs. Co-occurrence edges use JOINT and contrastive evidence to name same-layer relations; directed transcoder edges use sentence-grounded evidence to name source-to-target transformations.}
% \label{fig:autorelate}
% \end{figure}

\subsection{Why the views work together}

\subsubsection{From feature clutter to reusable scientific objects}

The filter, the hierarchy, the co-occurrence graph, and the mechanism graph solve different parts of the same practical problem. Without the filter, a researcher faces a flat inventory in which nuisance regions and semantically nearby regions are mixed together. Without the hierarchy, that retained universe is harder to navigate and harder to compress. Without co-occurrence, the retained universe has no corpus-level structure. Without the mechanism view, it has no sentence-conditioned account of how one concept family turns into another. The point is not that each view exists. The point is that each view answers a different question about the same semantic node universe.

\subsubsection{From corpus motifs to sentence-conditioned concept flow}

This shared referent makes the workflow scientifically coherent. A researcher can first define a retained universe because it is semantically relevant. They can then ask which motifs recur within that universe at multiple textual scales. Finally, on a selected sentence, they can ask which of those concepts are active and how they are being transformed across layers. Auto-related edge labels make both steps readable. The same set of concepts is now visible as a thematic region, as a corpus-level motif structure, and as an input-conditioned transformation graph.

% \begin{figure}[t]
% \centering
% \begin{minipage}[t]{0.48\linewidth}
% \centering
% \IfFileExists{figures/saegraph_webapp_hierarchy.png}{%
%     \includegraphics[width=\linewidth]{figures/saegraph_webapp_hierarchy.png}%
% }{%
%     \fbox{\parbox[c][40mm][c]{0.92\linewidth}{\centering
%     \textbf{Panel (a) placeholder}\\[2pt]
%     SAEGraph hierarchy browser screenshot\\
%     showing the abstraction tree interface}}
% }
% \\[2pt]
% {\footnotesize (a) Abstraction hierarchy interface.}
% \end{minipage}\hfill
% \begin{minipage}[t]{0.48\linewidth}
% \centering
% \IfFileExists{figures/saegraph_webapp_dualgraph_compressed.png}{%
%     \includegraphics[width=\linewidth]{figures/saegraph_webapp_dualgraph_compressed.png}%
% }{%
%     \fbox{\parbox[c][40mm][c]{0.92\linewidth}{\centering
%     \textbf{Panel (b) placeholder}\\[2pt]
%     SAEGraph dual-graph and compressed-view screenshot\\
%     showing linked corpus and mechanism views}}
% }
% \\[2pt]
% {\footnotesize (b) Dual-graph and compressed mechanism view.}
% \end{minipage}
% \caption{SAEGraph web application. Panel (a) shows the abstraction hierarchy used to organize the retained concept universe and define reusable grouped views. Panel (b) shows the linked downstream views: the graph interface for that universe together with the compressed mechanism view for a selected input. The web application mirrors the workflow of the paper and exposes the full processed corpora rather than a few isolated examples.}
% \label{fig:saegraph-webapp}
% \end{figure}

\subsection{Quantitative structure-recovery metrics}
\label{app:structure-metrics}

Table~\ref{tab:structure-metrics} complements the shared-coordinate density maps in Figure~\ref{fig:structure-main} and the full chapter atlas in Figure~\ref{fig:chapter-density-full}. The visual results show that the retained sentence graph recovers distinct chapter basins and finer subchapter refinement. The table shows the same pattern numerically. Paragraph- and sentence-level aggregation give the strongest alignment with textbook structure, while coarser chapter- and subchapter-level aggregation is less localized. This is consistent with the construction of the graph: sentence and paragraph aggregation preserve more of the local co-activation signal that organizes the retained universe into coherent semantic regions.

\begin{table}[t]
\centering
\small
\caption{Structure metrics for sentence co-occurrence graphs aggregated at different textual levels. Chapter and subchapter alignment report a scaled mutual-information score between graph partitions and dominant textbook labels; because this score is not bounded by 1, values above 1 are possible. Same-chapter mass is the fraction of total edge weight whose endpoints share a dominant chapter. Within/between distance is the mean distance between same-chapter nodes divided by the mean distance between different-chapter nodes; lower is better.}
\label{tab:structure-metrics}

\resizebox{\textwidth}{!}{%
\begin{tabular}{lcccc}
\hline
& \multicolumn{2}{c}{Structure Alignment} & \multicolumn{2}{c}{Graph Concentration} \\
Level
& \begin{tabular}[c]{@{}c@{}}Chapter align\\(scaled MI) $\uparrow$\end{tabular}
& \begin{tabular}[c]{@{}c@{}}Subchapter align\\(scaled MI) $\uparrow$\end{tabular}
& \begin{tabular}[c]{@{}c@{}}Same-chapter\\mass $\uparrow$\end{tabular}
& \begin{tabular}[c]{@{}c@{}}Within/Between\\distance $\downarrow$\end{tabular} \\
\hline
Sentence   & 2.005          & \textbf{0.849} & \textbf{0.870} & 0.828          \\
Paragraph  & \textbf{2.190} & 0.836          & 0.811          & \textbf{0.781} \\
Subchapter & 1.109          & 0.226          & 0.575          & 0.849          \\
Chapter    & 0.995          & 0.247          & 0.230          & 0.983          \\
\hline
\end{tabular}%
}
\end{table}

% \section{AI Usage}
% AI was used to edit the manuscript and for coding assistance.

\end{document}